\documentclass[lettersize,journal]{IEEEtran}
\usepackage{amsmath,amsfonts,amssymb}
\usepackage[ruled,vlined,linesnumbered]{algorithm2e}
\usepackage{tikz}
\usetikzlibrary{positioning}
\usepackage{array}
\usepackage[caption=false,font=normalsize,labelfont=sf,textfont=sf]{subfig}
\usepackage{textcomp}
\usepackage{stfloats}
\usepackage{url}
\usepackage{verbatim}
\usepackage{graphicx}
\usepackage{cite}
\usepackage{booktabs}
\usepackage{multirow}

\usepackage{tabularx}

\begin{document}

\title{PARM: Pipeline-Adapted Reward Model}

\author{
Xingyu Fan,~\IEEEmembership{Graduate Student Member,~IEEE},~%
Wei Shao$^{*\,\ddagger}$,~\IEEEmembership{Member,~IEEE},~%
Jiacheng Liu$^{\dagger}$,~\IEEEmembership{Member,~IEEE},~\\%
Linqi Song$^{\dagger}$,~\IEEEmembership{Member,~IEEE},~%
Pheng Ann Heng,~\IEEEmembership{Senior Member,~IEEE},%
\thanks{*~Co-first author. $\dagger$~Co-corresponding authors. $\ddagger$~Project leader.}
}



\maketitle

\begin{abstract}
Reward models (RMs) have become central to aligning large language models (LLMs) with human preferences, powering both reinforcement learning from human feedback (RLHF) and more effective decoding strategies. While most prior work focuses on reward models for single model's text generation, the growing complexity of real-world applications has led to the adoption of multi-stage or pipeline tasks, where LLMs serve as components in a sequential process. However, how to design and train reward models that can effectively guide such pipeline tasks remains underexplored. 
In this work, we systematically investigate the role of reward models in LLM-based pipelines, focusing on code generation for combinatorial optimization problems as a representative task. We construct a novel pipeline framework that integrates a reward model into both the formulation and solution generation stages, and we analyze the challenges (such as 
inconsistency between rewards from reward models and results from pipelines) that arise when extending reward guidance beyond single-step generation. 
To address these challenges, we propose the Pipeline-Adapted Reward Model (PARM), a training method that leverages pipeline-specific data and direct preference optimization to more accurately reflect downstream execution feedback. While the stage-wise principles underlying PARM can in principle extend to general $k$-stage pipelines, this paper presents and evaluates a two-stage instantiation (formulation $\rightarrow$ code generation) as a representative and rigorous proof-of-concept. We conduct extensive experiments on four public optimization benchmarks, evaluating execution rate and solving accuracy, and compare PARM with baseline reward models and sampling-based methods. We further include a supplementary cross-domain experiment on GSM8K to assess transfer beyond optimization. Our results demonstrate that PARM consistently improves pipeline output quality and stability across all datasets. Further analysis provides new insights into the interplay between reward modeling and multi-stage LLM reasoning.
This work highlights the importance of reward model adaptation in complex pipelines and provides a practical framework for their integration, paving the way for more robust LLM applications in real-world, multi-stage tasks.
\end{abstract}


\section{Introduction}\label{intro}
\IEEEPARstart{R}{ecently}, reward models have garnered significant attention for their central role in aligning large language models (LLMs) with human preferences, especially in reasoning-centric tasks. Reward models are now integral to both reinforcement learning from human feedback (RLHF)~\cite{ouyang2022training, luong2024reft} and advanced decoding strategies, enabling LLMs to generate outputs that are not only syntactically plausible but also semantically aligned with desired objectives. As scoring models, reward models have been widely adopted to evaluate and select among candidate outputs from LLMs, forming the backbone of techniques such as Best-of-N (BoN) sampling, Beam Search, and Monte Carlo Tree Search (MCTS). These strategies, by integrating reward signals, consistently yield outputs of higher factual accuracy, coherence, and task relevance.
For instance, Best-of-N sampling has demonstrated remarkable effectiveness in open-domain tasks like question answering and web search~\cite{nakano2021webgpt, stiennon2020learning}, offering a simple yet powerful way to select high-quality responses. In machine translation, Beam Search—often paired with modified scoring functions—remains a standard technique for enhancing translation quality~\cite{wu2016google, murray2018correcting}. More recently, the use of MCTS with policy-driven LLMs has enabled the autonomous generation of high-quality training data, bypassing the need for exhaustive human annotation~\cite{zhang2024rest}. Collectively, these advances have established reward models as indispensable tools for optimizing LLM outputs in a diverse range of single-stage language tasks.

At the same time, the rapid progress of LLMs has catalyzed their adoption in more complex, process-driven scenarios. In contemporary applications such as automated programming, multi-step problem solving, and workflow orchestration, LLMs are increasingly deployed as modular components within pipeline frameworks~\cite{gao2023pal, zhousolving}. Each stage of such a pipeline may address a distinct subtask—such as problem formulation, code synthesis, or solution validation—with the overall system performance depending on the coordinated interaction of these stages. 

While these pipeline-based systems hold great promise for scaling LLM capabilities to tackle real-world, multi-stage problems, they also introduce new challenges for quality control. In practice, current LLM-based pipelines often fall short of delivering the reliability and output quality required for practical deployment, due to compounding errors and the lack of holistic evaluation signals. Drawing on the proven success of reward models in single-model settings, a natural question arises: \textbf{can reward models be leveraged to guide and enhance the output of entire LLM-driven pipelines?} However, despite their potential, the application of reward models in pipeline frameworks remains largely unexplored. Existing research predominantly focuses on reward-guided optimization of individual model outputs, neglecting the intricate dependencies and cumulative effects present in multi-stage pipelines. This gap highlights an urgent need to systematically investigate reward model design, training, and integration strategies tailored for pipeline-based LLM systems, in order to unlock their full potential in complex, process-oriented applications.

\begin{figure}[!t]
    \centering
    \includegraphics[width=\columnwidth]{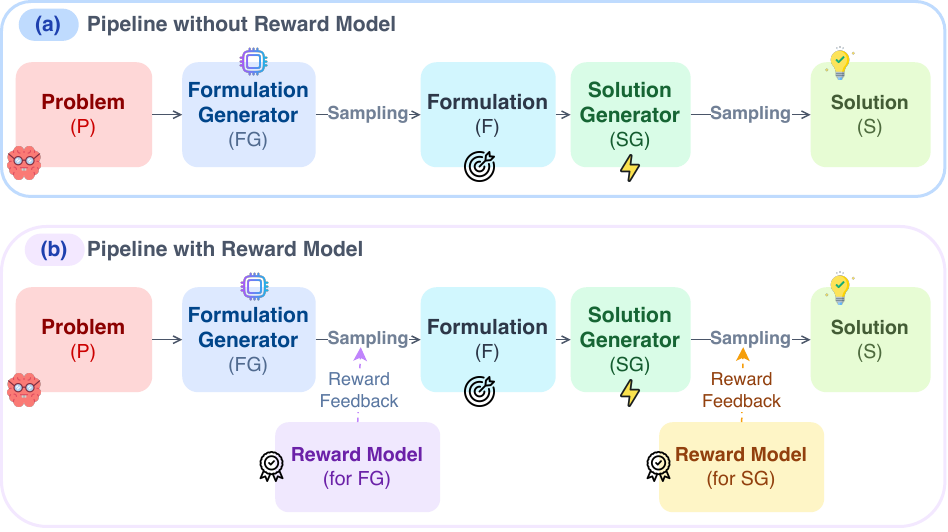}
    \caption{Comparison of the pipeline without reward model guidance (a) and with reward model guidance (b). In the latter, the generator produces multiple candidates, and the reward model scores these outputs. The final generation is sampled based on the reward scores.}
    \label{pipeline_comparison}
\end{figure}

\begin{figure*}[!t]
    \centering
    \includegraphics[width=\textwidth]{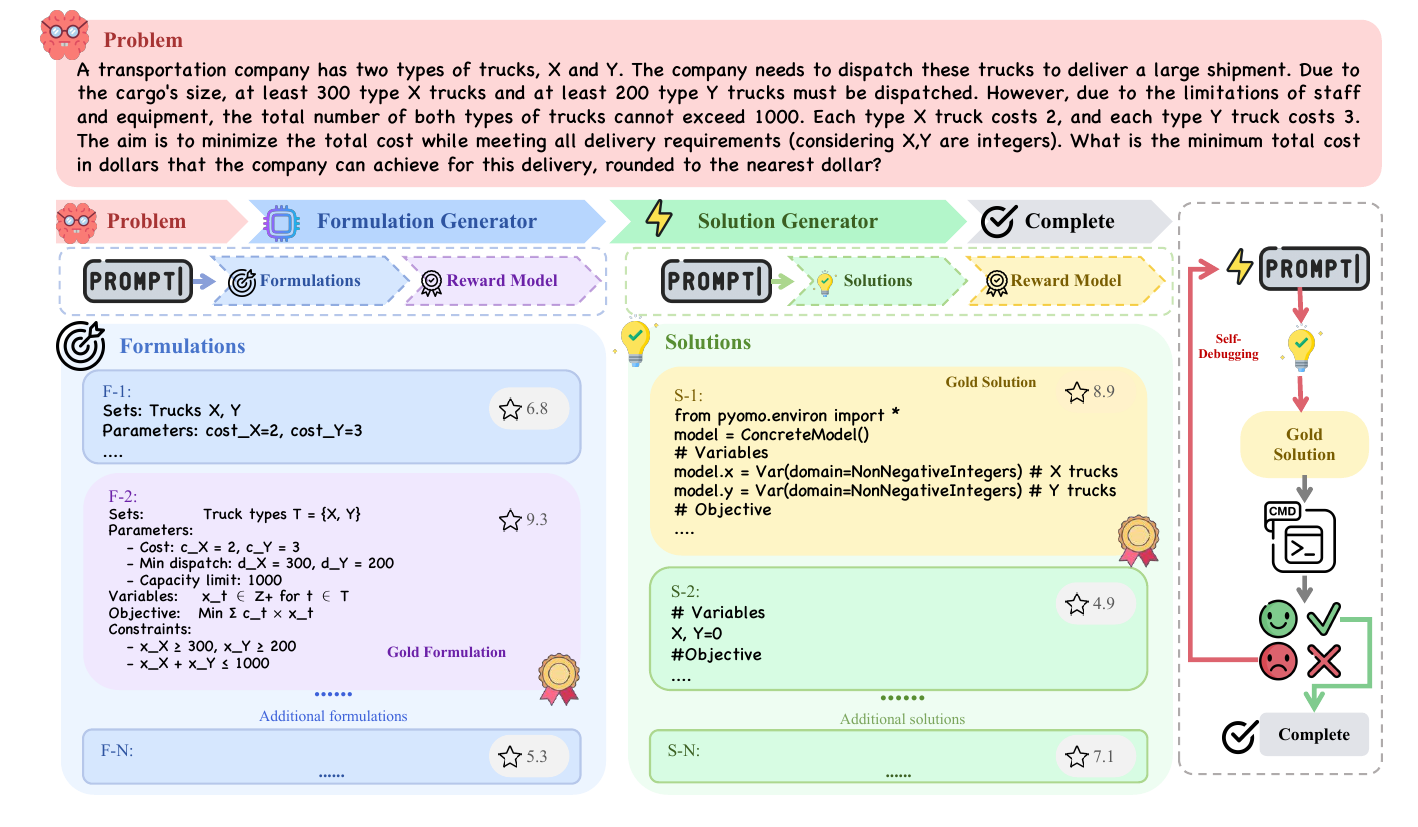}
    \caption{Illustration of the proposed pipeline given a concrete optimization problem. The figure shows the entire workflow, starting from a problem statement (top left), through the automatic generation of multiple problem formulations and solution candidates by large language models (LLMs), to their evaluation and selection using reward models. Example outputs at each stage are provided, including candidate formulations and solutions, as well as the final selected (gold) formulation and solution.}
    \label{example_pipeline}
\end{figure*}

To concretely investigate the role of reward models in pipeline frameworks, this paper takes code generation for mathematical optimization problems as a representative case study~\cite{tang2024orlm, jiang2024llmopt, xiao2023chain}. In our setting, the pipeline receives a problem described in natural language as input and decomposes the solution process into two sequential stages. The first stage focuses on transforming the problem statement into a precise mathematical formulation, while the second stage generates executable Python code to solve the formulated problem. Each stage is handled by a dedicated LLM component, and the overall output quality depends on the coordinated success of both stages.

Figure~\ref{pipeline_comparison} visually contrasts two pipeline variants: one operating without reward model guidance, and one where a reward model evaluates and selects among multiple candidate solutions at each stage. Figure~\ref{example_pipeline} further illustrates the end-to-end process with a concrete example. While we use mathematical optimization as a case study—such as problems from operations research (e.g., scheduling or resource allocation), the pipeline-adapted reward modeling principle is not inherently limited to this domain. This modular structure highlights the dependencies between each stage and the importance of effective coordination for achieving high-quality end-to-end solutions.

While reward models have been shown to enhance output quality within individual pipeline stages, their effectiveness in improving the overall pipeline output is not guaranteed. A key challenge lies in the fact that existing reward models are typically trained to assess the quality of isolated component outputs, without considering their downstream impact on the final solution. This can result in situations where a stage output receives a high reward score, but is nonetheless suboptimal for the global task. 
For example, a mathematically correct but overly complicated formulation may score highly in the first stage, yet lead to code generation failure or runtime errors in the subsequent stage—ultimately degrading the pipeline’s end-to-end performance.

To address these challenges, we propose a simple and effective framework that adapts reward models for pipeline-based problem solving. Unlike traditional methods that assign rewards to individual component outputs, our approach uses feedback from the final pipeline result. This allows the reward model to learn how early-stage decisions—such as problem formulation or intermediate solutions—affect the overall outcome. By focusing on the end-to-end performance, the reward model can better guide each stage of generation and selection towards the true task objective. This pipeline-adapted strategy not only bridges the gap between local and global evaluation, but may in principle extend to other complex, multi-stage tasks where coordinated optimization is essential, although this paper empirically validates only a two-stage instantiation.

We systematically evaluate our approach on four diverse pipeline benchmarks. Experimental results demonstrate that our pipeline-adapted reward model yields consistent improvements in end-to-end output quality compared to traditional stage-wise reward models. Furthermore, our analysis provides key insights into reward model behaviors, guidance mechanisms, and their integration with non-training-based pipeline enhancements. We highlight code generation for mathematical optimization as one representative use case.

In summary, the main contributions of this work are as follows:
\begin{itemize}
  \item
  We introduce a pipeline-adapted reward-modeling framework for multi-stage pipeline systems and study it empirically in a two-stage setting.
  \item
  We propose a simple and effective training method for developing pipeline-adapted reward models, which leverage feedback from the final pipeline output to overcome the limitations of stage-wise reward modeling.
  \item
  We comprehensively evaluate our approach on four pipeline benchmarks, demonstrating consistent improvements in output quality. 
  
  \item we provide in-depth analyses of reward model behaviors, guidance mechanisms, and their compatibility with training-free pipeline enhancement techniques.
\end{itemize}

\section{Related Work}
\subsection{Pipeline-based LLM Systems}
The pipeline paradigm, long established in classical deep learning~\cite{gulccehre2016knowledge, glasmachers2017limits, chang2020assessing}, has become increasingly prominent in the era of large language models (LLMs)~\cite{wies2022sub, liu2021generated}. In this framework, a complex problem is decomposed into a sequence of subtasks, each addressed by a dedicated module or operator. Historically, such pipelines consisted of a series of hand-crafted functions or lightweight neural networks, each responsible for a specific transformation or analysis step. With the advent of LLMs, pipeline systems are now constructed by chaining multiple LLM components, each specializing in a stage such as data interpretation, reasoning, generation, or verification.

Recent research demonstrates that pipeline-based LLM systems excel at handling multi-step reasoning, program synthesis, code generation, and other structured tasks where modular decomposition improves interpretability and manageability~\cite{gao2023pal, zhousolving, xiao2023chain}. These systems often assign different LLMs or specialized prompting strategies to distinct stages, enabling more effective use of model capacity and task-specific adaptation. For example, in program synthesis, an LLM pipeline may first translate a natural language specification into a formal representation, then generate executable code, and finally validate the output.

Conceptually, pipeline-based LLM systems can be viewed as a subset of more general multi-agent or tool-augmented frameworks, where multiple intelligent components collaborate to solve complex problems~\cite{chan2023chateval, li2023api, liang2025automated}. In such settings, the boundaries between pipeline, workflow orchestration, and multi-agent systems become fluid: pipelines emphasize sequential, stage-wise processing, while agent-based and tool-augmented approaches may involve more dynamic, interactive, or parallel coordination among components. Workflow orchestration frameworks further generalize this idea by enabling conditional branching, feedback loops, and adaptive module selection based on intermediate results.

Despite their promise, pipeline-based LLM systems face several open challenges. Error propagation across pipeline stages, the lack of holistic supervision, and difficulties in jointly optimizing for end-to-end performance remain significant bottlenecks. While modularization improves interpretability and scalability, it also increases the complexity of coordination and quality control. Effective reward modeling and evaluation strategies for such systems are still underexplored, especially when compared to the extensive literature on single-stage LLM optimization.

\subsection{Reward Models for Process-oriented AI}
Reward models have become a fundamental component for supervising and selecting the outputs of large language models, especially in single-stage settings where the evaluation of output quality is straightforward~\cite{ouyang2022training}. For example, reward-guided sampling and reinforcement learning from human feedback (RLHF) have significantly improved factuality and user satisfaction in open-domain generation tasks.

In contrast, applying reward models to process-oriented AI systems,such as pipelines or multi-stage workflows, remains a largely unsolved challenge. In these systems, the final output quality depends on the coordinated performance of multiple interdependent components. Traditional outcome-based reward models are often trained to assess the output of a single stage, neglecting the impact of intermediate results on downstream stages and the end-to-end solution~\cite{wen2024rethinking}. On the other hand, process-based reward models require human-annotated feedback for each intermediate step, which makes them highly resource-intensive and difficult to scale~\cite{guo2025deepseek}. As a result, both approaches can suffer from high local scores that fail to translate into globally optimal outcomes, and face difficulties in capturing error propagation and complex dependencies across the pipeline.

Some recent work has begun to explore reward assignment and evaluation beyond single-stage scenarios. Approaches include distributing reward signals across agents in multi-agent systems, or deriving the reward from the final system output to better capture holistic performance~\cite{zhang2024rest, jiang2024enhancing}. However, systematic frameworks for designing, training, and benchmarking reward models tailored to process-oriented LLM systems are still lacking. Key challenges include credit assignment, reward sparsity, and the need for evaluation protocols that reflect the complexities of multi-stage workflows.

More recently, several works have pushed process reward modeling beyond final-outcome supervision. PAV~\cite{setlur2025pav}, VersaPRM~\cite{zeng2025versaprm}, and DG-PRM~\cite{yin2025dgprm} explore richer process-level supervision, while Beirami et al.~\cite{beirami2025bon} analyze the limits of pure Best-of-N outcome selection. These studies are adjacent to PARM in that they motivate finer-grained reward signals, but they do not study stage-wise reward adaptation for optimization pipelines.

\subsection{Improving LLM Output Quality}
Recent advances in improving LLM outputs have focused on two primary approaches: in-context learning~\cite{brown2020language, liu2023pre} and parameter-updating methods~\cite{luong2024reft,ouyang2022training}. In-context learning enhances generation through strategic prompting, exemplified by GPT-3~\cite{brown2020language} and Chain-of-Thought (CoT) prompting~\cite{wei2022chain}, which enable complex reasoning through intermediate steps. Parameter-updating approaches explicitly train model weights, with methods like RLHF~\cite{ouyang2022training} and ReFT~\cite{luong2024reft} leveraging reinforcement learning to align outputs with human preferences. As scoring models for evaluating LLM outputs, several decoding methods are also employed to generate higher-quality outputs. 

As scoring models for evaluating LLM outputs, several decoding methods are also widely adopted to generate higher-quality results. For instance, Best-of-N sampling has shown strong practical effectiveness~\cite{nakano2021webgpt, stiennon2020learning}, despite its simplicity and lack of additional training requirements. In machine translation, Beam Search remains a standard, with various modifications to the scoring function for improved results~\cite{wu2016google, murray2018correcting}. More recently, Monte Carlo Tree Search (MCTS) has been applied to policy LLMs to generate high-quality data without human-provided annotations~\cite{zhang2024rest}.

While effective, these methods incur substantial computational costs as models scale~\cite{liu2023pre}. Pipeline frameworks have emerged as an efficient alternative, decomposing complex tasks into specialized components. However, existing reward models focus solely on individual components, overlooking the quality of pipeline-level outputs.

\subsection{LLMs in Mathematical Optimization}
LLMs have demonstrated growing capability in mathematical optimization~\cite{gao2023pal, zhousolving, cobbe2021training}, evolving from direct computation~\cite{gao2023pal} to more sophisticated strategies. Recent research has expanded this line of work to more complex optimization tasks, as exemplified by competitions like NL4Opt~\cite{ramamonjison2023nl4opt}, which encourage the use of LLMs to extract, understand, and solve optimization problems described in natural language. Early approaches such as PAL~\cite{gao2023pal} and CSV~\cite{zhousolving} translated problem statements into executable code and incorporated self-verification to improve accuracy.

More recent work has emphasized structured decomposition for optimization tasks. Chain-of-Expert (CoE)~\cite{xiao2023chain} introduced specialized agents for distinct solution stages, while ORLM~\cite{tang2024orlm} proposed a three-element problem formulation. LLMOPT further developed a more structured five-element representation, decomposing problems before code generation and demonstrating that well-defined, often manually annotated, intermediate formulations can substantially enhance solution quality.

Concurrent with our work, recent optimization-oriented LLM systems have also explored structured decomposition and automated formulation. OR-LLM-Agent~\cite{zhang2025orllmagent} and AlphaOPT~\cite{kong2025alphaopt} use multi-stage workflows guided by solver or execution outcomes, while CAFA~\cite{deng2024cafa} and OptimAI~\cite{thind2025optimai} broaden the comparison to recent formulation and agentic optimization methods. These works improve generators or system design, whereas PARM focuses on adapting reward models for stage-wise selection.

Despite these advances, several limitations remain. Many approaches rely on proprietary LLM interfaces and manually curated datasets, which are costly to construct and annotate. Methods often depend on expensive manual data preparation, fine-tuning, and the computational overhead of large models such as GPT-4. Furthermore, current multi-step systems typically deploy a single large model for each pipeline stage, rather than specialized, smaller models for individual sub-tasks, raising concerns about scalability and efficiency.

\section{Methodology}\label{PARM}

\begin{figure*}[!t]
    \centering
    \includegraphics[width=\textwidth]{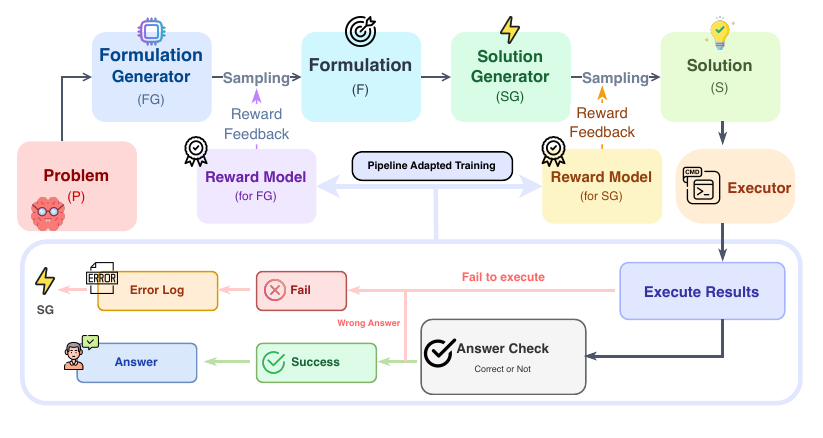}
    \caption{Overview of the proposed pipeline. The reward models evaluate both the formulation and solution generation stages, while execution feedback is utilized for pipeline-adapted training.}
    \label{framework}
\end{figure*}

As illustrated in Fig.~\ref{framework}, the PARM pipeline is composed of a Formulation Generator, a Solution Generator, and corresponding reward models for both stages. The Formulation Generator produces a mathematical formulation of the optimization problem in natural language, while the Solution Generator generates executable code based on the selected formulation. For each stage, multiple candidates are generated and subsequently scored by the respective reward models, with the highest-scoring outputs sampled as the final formulation and solution.

To ensure that the reward models effectively guide the overall pipeline output, we introduce a pipeline-adapted training (PAT) method. Specifically, task-level verification feedback from the final solution is leveraged as the supervision signal to adapt the reward models to the pipeline setting. The following sections detail each component and training methodology.

\subsection{Pipeline Framework}

Given an optimization problem $\mathrm{P}$, our pipeline leverages large language models (LLMs) and reward models to sequentially generate and select both the problem formulation and its solution. The overall workflow is summarized in Algorithm~\ref{alg:parm_pipeline}.

We denote the formulation generator as $\mathrm{LLM}_{\mathrm{F}}$, the solution generator as $\mathrm{LLM}_{\mathrm{S}}$, and their respective reward models as $\mathrm{RM}_{\mathrm{F}}$ and $\mathrm{RM}_{\mathrm{S}}$. $\mathrm{F}$ represents a candidate formulation and $\mathrm{S}$ a candidate solution. The process involves two main stages: (1) formulation generation and selection, and (2) solution generation and selection. For clarity, Algorithms~\ref{alg:parm_pipeline} and~\ref{alg:pipeline_train} present the two-stage instantiation used in our experiments; extending the same stage-wise selection and downstream-verification principles to additional stages is conceptually straightforward, but remains outside the empirical scope of this paper.

Specifically, given a problem $\mathrm{P}$, the pipeline first employs $\mathrm{LLM}_{\mathrm{F}}$ to sample a diverse set of candidate formulations $\{\mathrm{F}_i\}_{i=1}^{N_\mathrm{F}}$. Each formulation is then evaluated by the formulation reward model $\mathrm{RM}_{\mathrm{F}}$, which scores the candidates based on their downstream potential (e.g., executability or correctness of solutions they may yield). The formulation with the highest reward score is selected as the optimal formulation $\mathrm{F}_\mathrm{f}$.

Next, $\mathrm{F}_\mathrm{f}$ is passed to the solution generator $\mathrm{LLM}_{\mathrm{S}}$, which produces multiple solution candidates $\{\mathrm{S}_j\}_{j=1}^{N_\mathrm{S}}$. These candidate solutions are assessed by the solution reward model $\mathrm{RM}_{\mathrm{S}}$, and the top-ranked solution $\mathrm{S}_\mathrm{c}$ is chosen. 

To further enhance robustness and correctness, the pipeline incorporates a self-debugging loop. If the selected solution $\mathrm{S}_\mathrm{c}$ does not successfully solve the problem (as determined by automated correctness checks), error information is extracted and provided, alongside the original problem, to the solution generator. This feedback-driven mechanism enables $\mathrm{LLM}_{\mathrm{S}}$ to iteratively refine and resample solutions, guided by reward model scoring, until a correct solution is obtained or a retry threshold is reached.

\begin{algorithm}[t]
\caption{PARM Pipeline Workflow}
\label{alg:parm_pipeline}
\SetKwInOut{Input}{Input}
\SetKwInOut{Output}{Output}
\Input{Problem $\mathrm{P}$; Formulation generator $\mathrm{LLM}_{\mathrm{F}}$; Solution generator $\mathrm{LLM}_{\mathrm{S}}$; Formulation Reward Model $\mathrm{RM}_{\mathrm{F}}$; Solution Reward Model $\mathrm{RM}_{\mathrm{S}}$}
\Output{Final solution $\mathrm{S}_\mathrm{f}$}

Generate $\mathrm{N}_\mathrm{F}$ formulations: $\mathrm{F}_\mathrm{i} = \mathrm{LLM}_{\mathrm{F}}(P),~\mathrm{i}=1,\ldots,\mathrm{N}_\mathrm{F}$\;
Score each formulation with formulation-stage context: $\mathrm{r}_\mathrm{i}^\mathrm{F} = \mathrm{RM}_{\mathrm{F}}(P, \mathrm{F}_\mathrm{i})$\;
Select best formulation: $\mathrm{F}_\mathrm{f} = \arg\max_{\mathrm{F}_\mathrm{i}} \mathrm{r}_\mathrm{i}^\mathrm{F}$\;
Generate $\mathrm{N}_\mathrm{S}$ solutions: $\mathrm{S}_\mathrm{j} = \mathrm{LLM}_{\mathrm{S}}(\mathrm{F}_\mathrm{f}),~j=1,\ldots,\mathrm{N}_\mathrm{S}$\;
Score each solution with solution-stage context: $\mathrm{r}_\mathrm{j}^\mathrm{S} = \mathrm{RM}_{\mathrm{S}}(\mathrm{F}_\mathrm{f}, \mathrm{S}_\mathrm{j})$\;
Select best solution candidate: $\mathrm{S}_\mathrm{c} = \arg\max_{\mathrm{S}_\mathrm{j}} \mathrm{r}_\mathrm{j}^\mathrm{S}$\;

\While{not \textbf{IsCorrect}($\mathrm{S}_\mathrm{c}$, $P$) \textbf{and} retry limit not reached}{
    Execute $\mathrm{S}_\mathrm{c}$ and collect error information\;
    Feed error feedback and $P$ to self-debugging generator: \\
    \hspace{1.5em} Generate $\mathrm{N}_\mathrm{D}$ debugged solutions: $\mathrm{S}_\mathrm{d}^{(k)} = \mathrm{LLM}_{\mathrm{S}}(\mathrm{P}, \text{error info}),~k=1,\ldots,\mathrm{N}_\mathrm{D}$\;
    Score each debugged solution with solution-stage context: $\mathrm{r}_\mathrm{d}^{(k)} = \mathrm{RM}_{\mathrm{S}}(\mathrm{F}_\mathrm{f}, \mathrm{S}_\mathrm{d}^{(k)})$\;
    Select best debugged solution: $\mathrm{S}_\mathrm{c} = \arg\max_{\mathrm{S}_\mathrm{d}^{(k)}} \mathrm{r}_\mathrm{d}^{(k)}$\;
}

\Return $\mathrm{S}_\mathrm{c}$\;
\end{algorithm}

\textbf{Key features of our pipeline include:} (i) modularity, enabling independent improvements or substitutions of generators and reward models at each stage; (ii) the use of reward models to align both formulation and solution generation with downstream task success; and (iii) the integration of error-driven self-debugging to improve final solution quality. This framework also seamlessly supports pipeline-adapted training strategies, where reward models and generators can be iteratively improved based on pipeline feedback---an aspect we will discuss in detail in subsequent sections.

Overall, this two-stage, reward-driven approach enables systematic exploration of both problem formulations and solutions, leveraging the complementary strengths of LLMs and learned reward signals to automate and optimize the full problem-solving process.

\subsection{Generator}
The Generator module in our pipeline is responsible for producing both the mathematical formulation and the executable solution for a given optimization problem. Although we use mathematical optimization problems from operations research as a representative scenario, our approach is designed to be general and adaptable to a broad range of problem types. The modular generator structure, together with reward-based selection, constitutes the core innovation of our pipeline architecture, independent of any specific task domain.

Given a problem statement $P$ (see Fig.~\ref{example_pipeline} for an illustration), the pipeline first invokes the \emph{Formulation Generator} to produce candidate mathematical formulations. To guide this process, we design domain-agnostic prompts and instructions (details in Appendix: Prompts Design) for the LLM, focusing on decomposing $P$ into a structured "five-element formulation" (e.g., sets, parameters, variables, objectives, and constraints)~\cite{jiang2024llmopt, ramamonjison-etal-2022-augmenting}. This structured decomposition facilitates both interpretability and downstream solution generation. Details of the prompt design can be found in Appendix. 

For each problem, we employ a \textbf{best-of-$\mathrm{N}$ sampling} strategy, randomly sampling $N=32$ candidate formulations from the LLM. Each candidate is independently scored by a formulation reward model, and the highest-scoring formulation is selected for the next stage. This reward-guided selection ensures that the most promising and well-structured formulation is used as the basis for solution generation.

Next, the selected formulation is passed to the \emph{Solution Generator}, which is also realized by an LLM but with a different set of prompts tailored for generating executable Python code. These prompts are designed to instruct the model to solve the optimization problem using appropriate code structures and libraries. Again, we adopt a best-of-$\mathrm{N}$ sampling strategy ($N=32$), producing multiple candidate solutions. Each solution candidate is scored by a solution reward model, and the top candidate is selected as the final output to be executed.

This two-stage generator design, driven by reward models and best-of-$\mathrm{N}$ sampling, offers several advantages:
\begin{itemize}
    \item \textbf{Modularity and extensibility}: Each generator can be individually improved or adapted to other domains without altering the overall pipeline.
    \item \textbf{Increased robustness}: Sampling a diverse set of candidates and applying reward-based selection significantly improves the likelihood of obtaining both valid formulations and correct executable solutions.
    \item \textbf{Pipeline synergy}: The design ensures that only high-quality intermediate results are passed downstream, reducing error propagation and enhancing end-to-end performance.
\end{itemize}

It is important to note that the focus of our work is not on achieving domain-specific state-of-the-art results, but rather on demonstrating how a reward-guided, multi-stage generation pipeline can systematically improve the reliability and interpretability of LLM-based problem solving. Additional details, including prompt templates and concrete examples for both generators, are provided in the Appendix.

\subsection{Pipeline Reward Models}
\textbf{Design Motivation:}  
The motivation for introducing reward models in our pipeline stems from the desire for modular, efficient, and scalable multi-stage generation. Instead of relying on a single large model for end-to-end task completion or expensive fine-tuning with large annotated datasets, we decompose the pipeline into phases, with each generator paired with a lightweight reward model that specializes in scoring and selecting candidates for that phase. This modular approach reduces annotation cost and enables the reward models to evolve using automatically collected feedback signals from the pipeline itself.

\textbf{Reward Model Evaluation Criteria:}  
For both the formulation and solution stages, the reward models are designed to evaluate the quality of candidates by jointly considering the input prompt and the generated candidate. We include task-specific examples in the prompt to help the reward model learn relevant evaluation cues. For formulations, the reward model is implicitly guided by mathematical correctness and structural completeness; for solutions, the model is further guided by code correctness and execution feasibility.  
To quantify final solution quality, we use two metrics:  
\begin{itemize}
    \item \textbf{Execution Rate (ER):} the proportion of generated code solutions that execute without error and produce valid outputs.
    \item \textbf{Solving Accuracy (SA):} the proportion of executed solutions whose optimal values match any provided ground truth.
\end{itemize}
These metrics also inform the construction of training data for the reward models.

\textbf{Input, Output, and Cooperation:}  
During inference, each reward model receives as input a prompt-candidate pair from the corresponding generator and outputs a scalar score for each candidate. The generator then selects the highest-scoring candidate based on these scores. The reward models and generators are thus tightly coupled: the generator produces a diverse set of candidates, and the reward model ensures that only the most promising candidate advances to the next stage.

\begin{figure}[!t]
    \centering
    \includegraphics[width=\columnwidth]{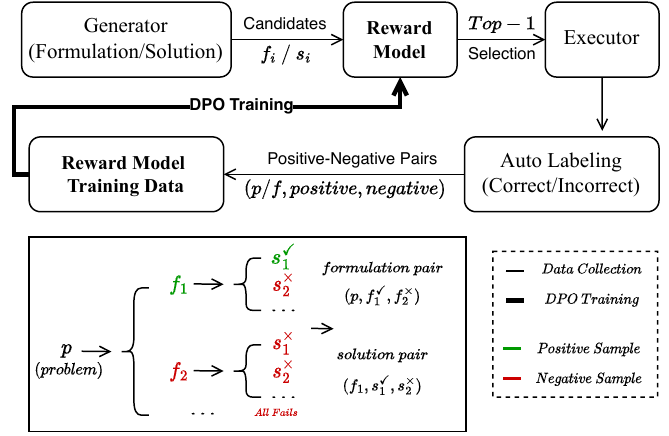}
    \caption{Reward signal flow and automated data collection for reward model training. Below shows the data collection paradigm.}
    \label{fig:reward_flow}
\end{figure}

\textbf{Automated Reward Signal Collection:}
A key innovation of our approach is the fully automated collection of reward model training data. As shown in Fig.~\ref{fig:reward_flow}, the pipeline labels candidates using the task-level verification criterion introduced below. For solution reward model training, a code sample is treated as positive if it satisfies the verification criterion and negative otherwise. For formulation reward model training, a formulation is treated as positive if at least one downstream solution is verified, and negative if no downstream solution is verified.
This process enables automatic construction of reward-model supervision without manual annotation.

In summary, our reward models enable the pipeline to improve its selection ability by leveraging task-level verification feedback while eliminating the need for manual labeling. This reward-driven modular selection is a core mechanism of our pipeline's adaptability and robustness.

\subsection{Pipeline-adapted Training Algorithm}\label{reward_training}
Conventional reward models for LLMs are typically trained on general human preference data and are not tailored to judge the specific outputs of modular, multi-stage pipelines. As a result, their score predictions may not accurately reflect the correctness or utility of pipeline outputs, especially when success is determined by downstream execution or mathematical correctness. To address this, we propose a \emph{pipeline-adapted training} approach, in which the reward model is optimized to score pipeline outputs (formulations or code) based on their actual effectiveness in the entire pipeline context.

\textbf{Cross-Stage Reward Formulation.}\quad
A key theoretical contribution of PARM is the definition of \emph{cross-stage utility functions} as the supervision target for reward model training. Unlike conventional reward models that score a local output $y$ independently of downstream effects, PARM defines each stage's reward in terms of end-to-end pipeline success.

Formally, let $\mathbf{1}[\cdot]$ denote the indicator function and let $\mathrm{Verify}(P, F_i, S_{ij})$ denote whether solution $S_{ij}$, generated from formulation $F_i$ for problem $P$, satisfies the task-level verification criterion used in our pipeline (e.g., successful execution and, when available, correctness against a reference answer or optimal value). The cross-stage utility for the formulation stage is:
\begin{align}
    R_\mathrm{F}(P,\, F_i) = \mathbf{1}\!\left[\exists\, j:\; \mathrm{Verify}(P, F_i, S_{ij}) = 1 \right],
\end{align}
and for the solution stage:
\begin{align}
    R_\mathrm{S}(P,\, F_i,\, S_{ij}) = \mathbf{1}\!\left[\mathrm{Verify}(P, F_i, S_{ij}) = 1\right].
\end{align}
Critically, $R_\mathrm{F}$ is not directly observable from $F_i$ alone---it requires executing the downstream pipeline and checking the resulting task-level outcome. This distinguishes PARM's supervision signal from standard local reward modeling.

Given these utilities, the ideal stage-wise selection would jointly maximize:
\begin{align}
    (F^*, S^*) = \arg\max_{F_i,\, S_{ij}}\; \mathbb{P}\!\left(\mathrm{Verify}(P, F_i, S_{ij}) = 1\right),
\end{align}
which is intractable to compute directly. PARM approximates this via the learned reward models:
\begin{align}
    F^* = \arg\max_i\; r_{\theta_\mathrm{F}}(P, F_i), \quad S^* = \arg\max_j\; r_{\theta_\mathrm{S}}(x_\mathrm{S}(F^*), S_j).
\end{align}
where $x_\mathrm{S}(F^*)$ denotes the solution-stage context constructed from the selected formulation (and any accompanying prompt information). The PARM training objective is designed to make these stage-wise scores correlate with downstream success, making this greedy approximation effective in practice.

\textbf{Automatic Construction of Training Data.}\quad
A key advantage of our pipeline is the ability to automatically generate high-quality preference data without human intervention. For every input problem, the pipeline generates multiple candidate formulations and solutions. By verifying these candidates against the task-level criterion described above, we can directly identify which outputs should be treated as positive samples and which should be treated as negative samples, thus constructing preference pairs for reward model training (the automatic labeling process is illustrated in Fig.~\ref{fig:reward_flow}):
\begin{itemize}
    \item \textbf{Formulation preference pairs:} For a given problem, a \emph{chosen} formulation is one that leads to at least one downstream solution satisfying the task-level verification criterion ($R_\mathrm{F}=1$); a \emph{rejected} formulation yields no verified downstream solution ($R_\mathrm{F}=0$).
    \item \textbf{Code preference pairs:} For a given formulation, a \emph{chosen} code sample satisfies the task-level verification criterion ($R_\mathrm{S}=1$), while a \emph{rejected} code sample does not ($R_\mathrm{S}=0$).
\end{itemize}

\textbf{Optimization with DPO.}\quad

To train our reward models from pairwise preference data, we employ the DPO objective~\cite{rafailov2023direct} in a stage-specific preference-learning setting. Unlike RLHF methods such as PPO, which fine-tune a policy model online, our training keeps the generators fixed and optimizes the stage-specific scorers from automatically constructed preference pairs.

Given the formulation-stage dataset $\mathcal{D}_\mathrm{F} = \{(P, F^+, F^-)\}$ and the solution-stage dataset $\mathcal{D}_\mathrm{S} = \{(x_\mathrm{S}, S^+, S^-)\}$, we instantiate the objective separately for the formulation and solution reward models:

\begin{scriptsize}
\begin{align}
\mathcal{L}_{\mathrm{F}}(\theta_\mathrm{F}) =\ & - \mathbb{E}_{(P, F^+, F^-) \sim \mathcal{D}_\mathrm{F}} \left[
    \log \sigma \left( \frac{r_{\theta_\mathrm{F}}(P, F^+) - r_{\theta_\mathrm{F}}(P, F^-)}{\beta} \right)
\right], \\
\mathcal{L}_{\mathrm{S}}(\theta_\mathrm{S}) =\ & - \mathbb{E}_{(x_\mathrm{S}, S^+, S^-) \sim \mathcal{D}_\mathrm{S}} \left[
    \log \sigma \left( \frac{r_{\theta_\mathrm{S}}(x_\mathrm{S}, S^+) - r_{\theta_\mathrm{S}}(x_\mathrm{S}, S^-)}{\beta} \right)
\right].
\end{align}
\end{scriptsize}

Here, $\beta$ is a temperature hyperparameter that controls the sharpness of preference enforcement. Intuitively, minimizing these losses encourages the reward models to assign higher scores to candidates that are more likely to lead to verified downstream outcomes, and lower scores to those that are not. Crucially, the preference labels in our setting are induced by the cross-stage utility functions $R_\mathrm{F}$ and $R_\mathrm{S}$, and therefore by the task-level verification signal rather than by human or LLM judgments. This makes the learned reward signal directly aligned with end-to-end pipeline success under the evaluation protocol used in this work.

Because the generators remain fixed, the optimization focuses solely on fitting the stage-wise scorers to task-grounded preference pairs. In our experiments, this objective effectively aligns the reward signal with actual downstream success and improves the reliability of automatic candidate selection in the pipeline.

\textbf{Training Workflow.}\quad The overall pipeline-adapted training workflow is summarized as Algorithm~\ref{alg:pipeline_train}.

\begin{algorithm}[t]
\caption{Pipeline-adapted Reward Model Training}
\label{alg:pipeline_train}
\SetKwInOut{Input}{Input}
\SetKwInOut{Output}{Output}
\Input{Problem set $\mathcal{P}$; Formulation generator $\mathrm{LLM}_{\mathrm{F}}$; Solution generator $\mathrm{LLM}_{\mathrm{S}}$}
\Output{Trained reward models $\mathrm{r}_{\theta_\mathrm{F}}$ and $\mathrm{r}_{\theta_\mathrm{S}}$}
\BlankLine
Initialize preference datasets $\mathcal{D}_\mathrm{F}$ and $\mathcal{D}_\mathrm{S}$\;
\For{problem $p \in \mathcal{P}$}{
    Generate $\mathrm{N}$ candidate formulations $\{\mathrm{F}_\mathrm{i}\}$\;
    \For{each $\mathrm{F}_\mathrm{i}$}{
        Generate $\mathrm{N}$ candidate solutions $\{\mathrm{S}_{\mathrm{i}\mathrm{j}}\}$\;
        Evaluate each $\mathrm{S}_{\mathrm{i}\mathrm{j}}$ with the task-level verification criterion and record the outcome\;
    }
    Add formulation preference pairs to $\mathcal{D}_\mathrm{F}$ based on whether any downstream solution is verified\;
    Add code preference pairs to $\mathcal{D}_\mathrm{S}$ based on the verification outcomes of candidate solutions\;
}
Train $\mathrm{r}_{\theta_\mathrm{F}}$ on $\mathcal{D}_\mathrm{F}$ and $\mathrm{r}_{\theta_\mathrm{S}}$ on $\mathcal{D}_\mathrm{S}$ using the DPO objective\;
\Return{$\mathrm{r}_{\theta_\mathrm{F}}, \mathrm{r}_{\theta_\mathrm{S}}$}
\end{algorithm}

\textbf{Relationship to RLHF and DPO.}\quad
Standard RLHF~\cite{ouyang2022training} trains a reward model on human preferences and then uses it---via PPO---to fine-tune the \emph{generator} (the LLM). DPO~\cite{rafailov2023direct} bypasses the explicit reward model and directly fine-tunes the generator with a contrastive objective. Both paradigms therefore target the \emph{generator}.

PARM keeps the generators fixed and instead applies a preference-learning objective to the \emph{reward models} that score and select stage outputs. It borrows the DPO pairwise loss for training the stage-specific selectors, but does \emph{not} update the generators. The two directions are complementary and composable: RLHF or policy-level DPO could align the generators, while PARM adapts the stage-wise reward models.

A practical advantage is the elimination of manual annotation. Because the labels are obtained automatically from the task-level verifier, the training signal is objective and reproducible. Compared with off-the-shelf reward models, the pipeline-adapted selectors learn cross-stage patterns that correlate with end-to-end success, enabling more reliable candidate selection.

\subsection{Self-Debugging Integration}

To further enhance the robustness and correctness of pipeline outputs, we incorporate a self-debugging mechanism into the solution generation stage (details in Appendix: Prompts Design). When a generated code candidate fails to compile or execute correctly, the associated compilation errors and diagnostic information, along with the original problem statement, are fed back into the solution generator in a self-debugging loop. The solution generator then attempts to revise and regenerate code based on this feedback, producing multiple new candidates. As in the initial generation phase, the reward model evaluates these candidates and selects the highest-scoring one for execution.

This process may iterate multiple times: if the selected correction still fails, further cycles of error feedback and code regeneration are performed until a valid solution is obtained or a predefined retry limit is reached. The final execution result is then compared against the ground-truth solution for the problem $P$, ensuring that only correct and executable code is accepted as the pipeline output.

By integrating self-debugging with reward-guided selection, our pipeline can dynamically correct common generation errors, significantly improving the reliability and success rate of automated solution synthesis without requiring manual intervention.

\section{Experiments}

\subsection{Overview}

To systematically evaluate the effectiveness of PARM in the studied setting and probe its transfer potential, we conduct our main empirical study on a diverse set of optimization datasets and complement it with a supplementary cross-domain experiment on GSM8K. In our evaluation, optimization serves as the primary case study for a two-stage pipeline, while GSM8K is included as a targeted transfer check beyond optimization. Our focus is not on achieving domain-specific state-of-the-art (SOTA) results, but on examining whether pipeline-adapted reward modeling improves end-to-end performance under this two-stage decomposition.

Table~\ref{statistic_datasets} summarizes the optimization datasets used in our main empirical study. These datasets, widely adopted in the literature, cover a broad spectrum of problem types and complexity levels, ranging from high-school-level mixed integer linear programming (MILP) to undergraduate-level complex optimization tasks in operations research. This diversity allows us to stress-test the core pipeline mechanism under various scenarios. A supplementary GSM8K experiment is introduced later in Section~\ref{sec:gsm8k} to test transfer beyond optimization.

\begin{table}[ht!]
    \centering
    \caption{Statistics of the optimization datasets used in the main empirical study.}
    \label{statistic_datasets}
    \begin{tabularx}{\columnwidth}{l|c|X}
        \hline
        \textbf{Dataset} & \textbf{Size} & \textbf{Description} \\ \hline
        IndustryOR~\cite{tang2024orlm}   & 100 & The first industrial dataset for optimization modeling, covering 13 industries and 5 problem types, including LP, MILP, and nonlinear programming. \\ \hline
        ComplexOR~\cite{xiao2023chain}   & 19  & Complex OR problems from academic and industrial sources, covering supply chain, scheduling, and logistics. \\ \hline
        NL4Opt~\cite{ramamonjison2023nl4opt} & 100 & Curated from the NL4Opt Competition, including LPWPs from sales, advertising, and investment, with exclusive target domains. \\ \hline
        NLP4LP~\cite{ahmaditeshnizioptimus}  & 65  & Derived from textbooks and lecture notes; includes facility location, network flow, scheduling, and portfolio management. \\ \hline
        Mamo Easy~\cite{huang2024mamo}     & 100 & High school-level MILP problems for basic optimization skills, part of the Mamo benchmark. \\ \hline
        Mamo Complex~\cite{huang2024mamo}  & 100 & Undergraduate-level LP and MILP problems for advanced learning and research, part of the Mamo benchmark. \\
        \hline
    \end{tabularx}
\end{table}

In all experiments, the pipeline structure remains the central research object. Shown in Table~\ref{tab:Expert_model}, we utilize several combinations of expert models (e.g., Qwen2.5-Series~\cite{yang2024qwen25mathtechnicalreportmathematical, hui2024qwen2}, Deepseek-Series~\cite{shao2024deepseekmath, deepseek-coder}) and reward models (e.g., Skywork-RM~\cite{skyworkopeno12024}, Qwen-PRM~\cite{prmlessons}). Notably, all expert models are used as-is, with no additional task-specific fine-tuning, highlighting the modularity and plug-and-play nature of the pipeline.

\begin{table}[!t]
\caption{Expert Models Used in the Pipeline Framework}
\label{tab:Expert_model}
\centering
\resizebox{\columnwidth}{!}{
\begin{tabular}{lcc}
\toprule
 & \textbf{Formulator} & \textbf{Coder} \\
\midrule
Qwen-Series      & Qwen2.5-Math-7B-Instruct & Qwen2.5-Coder-7B-Instruct \\
Deepseek-Series  & deepseek-math-7b-instruct & deepseek-coder-7b-instruct-v1.5 \\
\bottomrule
\end{tabular}
}
\end{table}

Our evaluation is organized to comprehensively assess the pipeline framework from multiple perspectives:

\begin{itemize}
    \item \textbf{Baseline and Component Analysis:} We compare PARM against strong LLM baselines and systematically analyze the individual and combined effects of key pipeline components, such as formulation decomposition, reward model selection, and the self-debugging mechanism.
    \item \textbf{Ablation Studies:} We further investigate the impact of configurable parameters (e.g., self-debugging iterations), different expert model and reward model pairings, and the scale of reward model training data.
    \item \textbf{Reward Model Evaluation:} We evaluate the contribution of reward models to final pipeline performance, illustrating their influence through quantitative metrics and planned effect curves.
\end{itemize}

Through this suite of experiments, our goal is to assess the robustness of PARM within the evaluated two-stage pipelines rather than to claim exhaustive validation across pipeline configurations. The optimization benchmarks form the main evaluation suite, while GSM8K serves as a supplementary out-of-domain validation. Together, they provide targeted evidence about the method in the settings studied here and suggest, rather than establish, broader applicability.

\subsection{Experimental Setup}
In this section, we detail the experimental settings for evaluating the PARM pipeline framework. Following the dataset summary in the previous subsection, we now describe the evaluation metrics, baseline configurations, implementation details, and training protocols for both the main pipeline and the reward models.

\textbf{Evaluation Metrics.}  
We employ two primary metrics to assess performance:
\begin{itemize}
    \item \textbf{Execution Rate (ER):} The proportion of generated solutions whose code can execute without errors and produce valid outputs.
    \item \textbf{Solving Accuracy (SA):} The proportion of evaluated solutions that match the task reference answer. For optimization tasks, this means that the executed solution's optimal value matches the provided ground-truth optimum; for GSM8K, it means that the final numerical answer exactly matches the reference answer.
\end{itemize}

\textbf{Baselines.}  
For a comprehensive comparison, we evaluate PARM against powerful large language model baselines, including GPT-4o and DeepSeek-v3. All baselines are assessed under the same two-step prompt protocol (generating a formulation, then generating solving code) to ensure fairness and consistency.

\textbf{Pipeline Configuration.}  
All experiments are implemented using the PyTorch framework, with vLLM employed for efficient LLM inference. The main pipeline evaluation is conducted on four NVIDIA A40 Tensor Core GPUs (each with 48 GB memory), while reward model training uses a single A40 GPU. Unless otherwise specified, the pipeline adopts the following default hyperparameters: a temperature of 0.3, a maximum generation length of 1280 tokens, 32 samples per step, one iteration for self-debugging, and a self-debugging sample size of 16.

\textbf{Reward Model Training.}  
The reward models are trained using Direct Preference Optimization (DPO), leveraging the Hugging Face TRL library. For efficient adaptation to large-scale models, we use LoRA (Low-Rank Adaptation) with the following default hyperparameters in the optimization experiments: LoRA rank $=128$, LoRA alpha $=64$, learning rate $=5.0 \times 10^{-7}$, number of epochs $=5$, beta $=0.1$, and evaluation split ratio $=0.1$. The supplementary GSM8K experiment follows the same overall setup, except that it uses 3 training epochs, as described in Section~\ref{sec:gsm8k}.
In line with the pipeline's automated design philosophy, preference data for reward model training is collected automatically from pipeline runs, requiring no manual annotation. This process is detailed in Section~\ref{reward_training}. The Mamo dataset (Easy and Complex subsets) is used to generate preference pairs for DPO training for the optimization experiments, and the corresponding optimization benchmarks are used as test datasets for evaluating reward model performance. We systematically vary both data difficulty and sample size in subsequent analyses.

\textbf{Reproducibility.}  
All code and experimental scripts are managed for full reproducibility. Further implementation details, including data preprocessing, model checkpoints, and prompt templates, are provided in the supplementary materials and will be released upon publication.

\subsection{Experimental Results}
\subsubsection{\textbf{Comparison with Baselines}}
To evaluate the effectiveness of the PARM pipeline mechanism, we compare its performance against two strong large language model baselines: GPT-4o~\cite{achiam2023gpt} and DeepSeek-v3~\cite{liu2024deepseek}. All methods are assessed under the same two-step protocol: first generating a five-element formulation, then generating solving code, to ensure a fair and consistent comparison.

\begin{figure}[!t]
    \centering
    \includegraphics[width=\columnwidth]{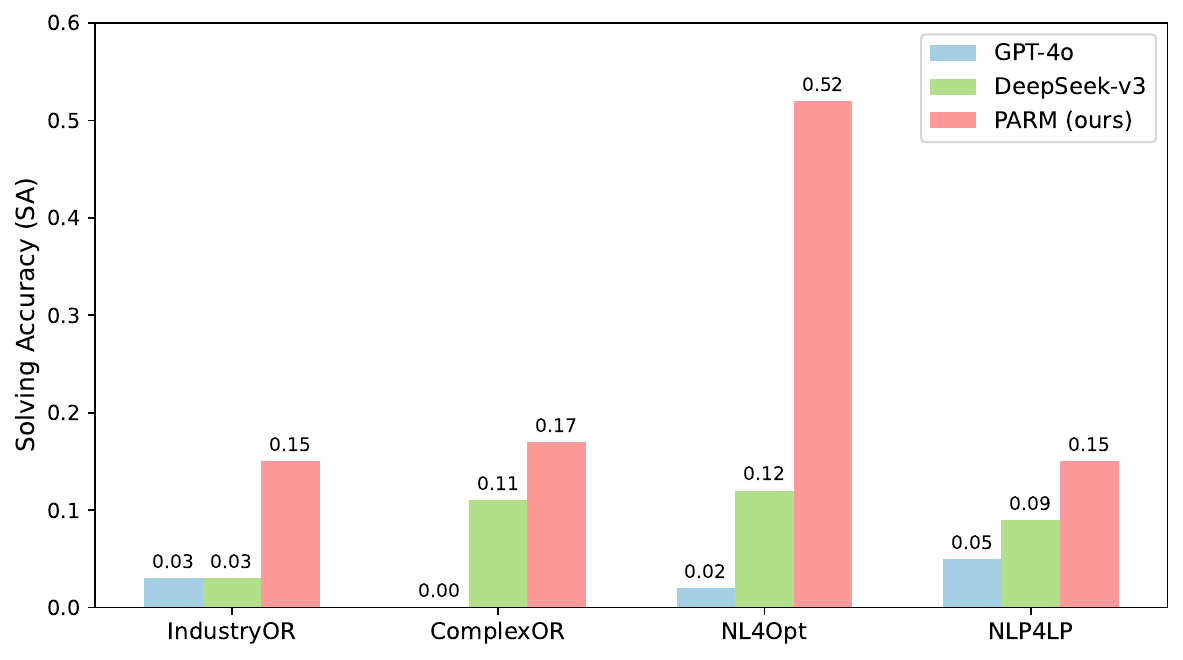}
    \caption{Comparison of the SA metric between PARM and Baselines across Four Datasets (1 ITERATION Self-Debugging)}
    \label{baseline_comparison}
\end{figure}

Figure~\ref{baseline_comparison} presents the Solving Accuracy (SA) across four representative datasets. As shown, PARM consistently outperforms both GPT-4o and DeepSeek-v3 on all benchmarks, despite utilizing expert models that have significantly fewer parameters than the foundations of these large models. This highlights the efficiency and effectiveness of the pipeline architecture, and more importantly, demonstrates that a modular pipeline of smaller expert models can surpass the direct application of much larger monolithic models.

It is important to emphasize that our goal is not to pursue task-specific SOTA in the optimization domain, but to study the behavior of PARM in a focused two-stage setting and test whether its benefits persist across several optimization benchmarks. The observed gains support the value of pipeline-adapted reward modeling in this setting, while broader validation across other pipeline structures remains future work.

\subsubsection{\textbf{Ablation Study on Pipeline Components}}
To systematically analyze the contribution and robustness of each component within the PARM pipeline, we conduct a series of ablation studies. Our goal is to highlight how each design choice—expert model selection, reward model configuration, self-debugging, and problem decomposition—affects the overall framework, and to demonstrate the pipeline's flexibility across various configurations.

\textbf{Expert Model Selection.}
Table~\ref{tab:Expert_model} and Table~\ref{main_experiment} compare the Qwen-Series and DeepSeek-Series as expert models within the pipeline. The results show that the Qwen-Series consistently outperforms the DeepSeek-Series across all datasets in both Execution Rate (ER) and Solving Accuracy (SA). This confirms that the pipeline structure is compatible with different expert models, while also showing that careful expert selection can further boost performance.

\textbf{Reward Model Configuration.}

As shown in Table~\ref{main_experiment}, we evaluate the impact of different reward model pairings (Skywork, Qwen-RM, and their DPO-trained variants). While this is not a strict architecture-only ablation, it provides a direct robustness check across reward-model backbones and adapted configurations. The pipeline demonstrates high robustness to these variations: both general-purpose reward models and pipeline-adapted ones yield competitive performance, and integrating reward models into the pipeline yields a significant improvement over pure sampling decoding.

\textbf{Self-Debugging Mechanism.}
Table~\ref{self_debug} presents the results of incorporating the self-debugging mechanism (limited to one iteration) into the pipeline. This component further improves both ER and SA on all datasets, underscoring the practical benefit of iterative refinement in complex reasoning tasks. The results validate the utility of self-debugging as a plug-and-play enhancement for the pipeline.

\begin{figure}[!t]
    \centering
    \includegraphics[width=\columnwidth]{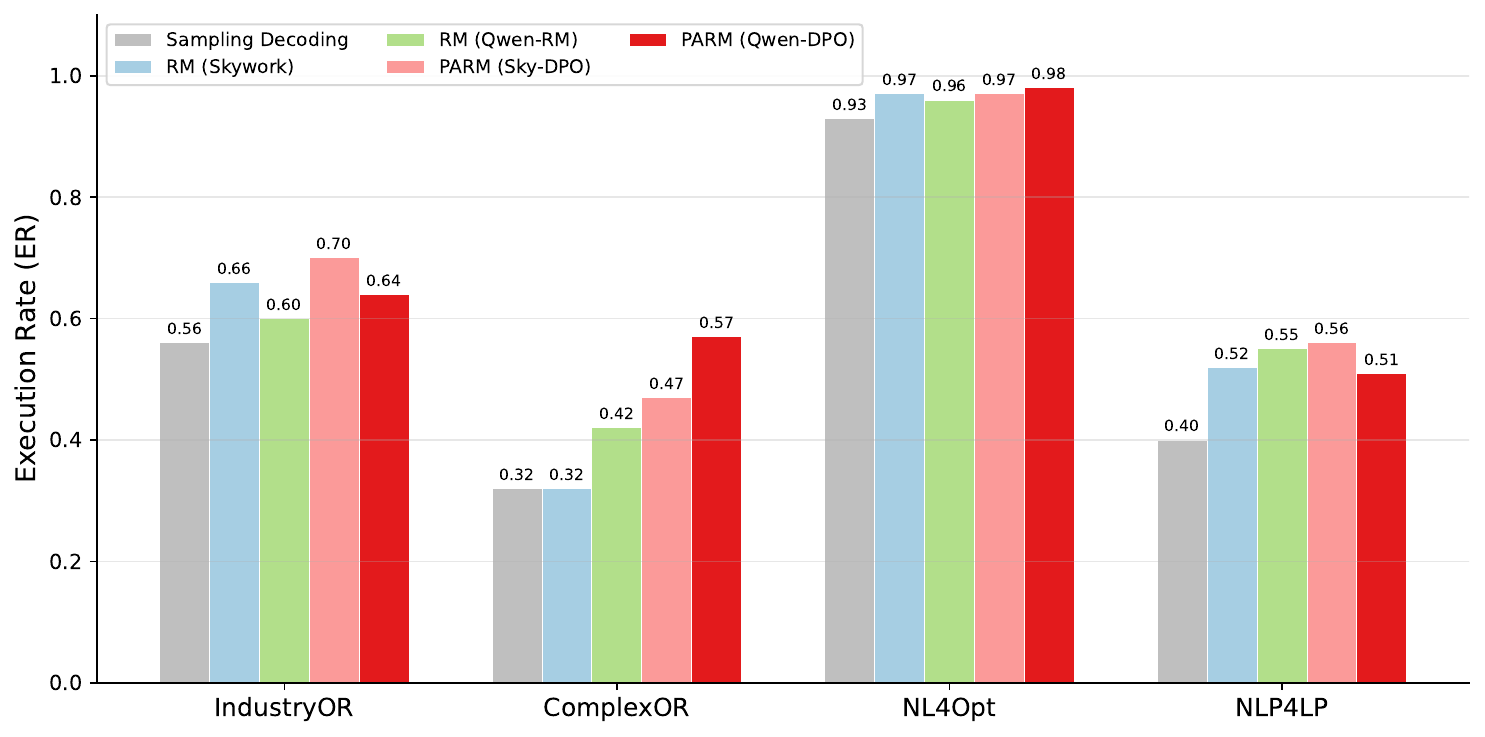}
    \caption{Execution Rate (ER) comparison across four datasets for different pipeline configurations (Qwen-Series, 1-iteration self-debugging).}
    \label{fig:er_barchart}
\end{figure}

\begin{figure}[!t]
    \centering
    \includegraphics[width=\columnwidth]{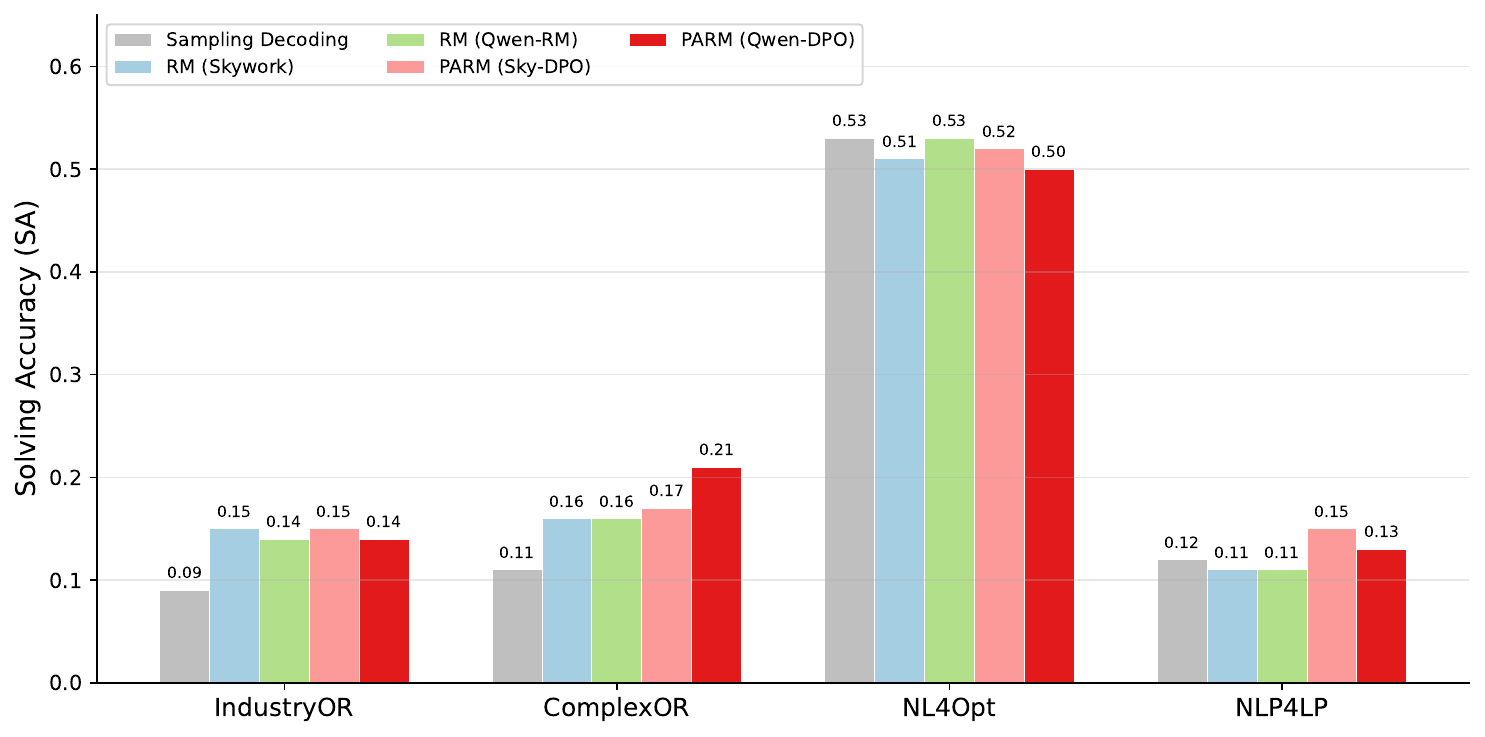}
    \caption{Solving Accuracy (SA) comparison across four datasets for different pipeline configurations (Qwen-Series, 1-iteration self-debugging).}
    \label{fig:sa_barchart}
\end{figure}

To complement the tabular results, Figure~\ref{fig:er_barchart} and Figure~\ref{fig:sa_barchart} present grouped bar charts of Execution Rate and Solving Accuracy, respectively, for the Qwen-Series pipeline configurations with 1-iteration self-debugging (corresponding to Table~\ref{self_debug}). These visualizations make the relative improvements across datasets and methods immediately apparent.

\begin{figure}[!t]
    \centering
    \includegraphics[width=\columnwidth]{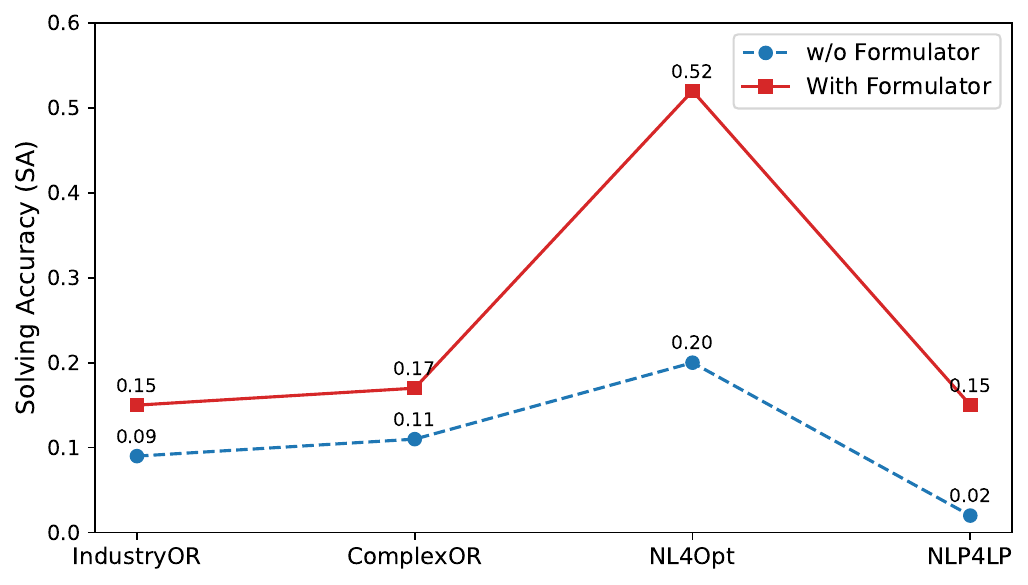}
    \caption{SA Metric of PARM With and Without Problem Decomposition (with / without Formulator)}
    \label{formulator_results}
\end{figure}

\textbf{Problem Decomposition (Formulator) Ablation.}
To assess the necessity of explicit problem decomposition, we compare the pipeline's performance with and without the Formulator (Figure~\ref{formulator_results}). Decomposing the problem into mathematical formulations markedly improves the solving accuracy, confirming the pivotal role of structured intermediate representations in optimization tasks and further supporting the pipeline's modular design.

Overall, these ablation studies demonstrate the flexibility, robustness, and extensibility of the PARM pipeline: it is able to integrate different expert models and reward models, and each component (reward model, self-debugging, formulator) brings measurable benefits to the framework as a whole. This validates our design philosophy of assembling lightweight, modular LLM-based agents for complex multi-stage problem solving.

\textbf{Per-Stage Reward Model Ablation.}
To isolate the independent \emph{inference-time} contribution of each stage-specific reward model, we conduct a controlled ablation experiment using the Qwen-Series pipeline (without self-debugging) on all four datasets. We use the default PARM DPO-adapted checkpoints in our runner: $\mathrm{RM}_\mathrm{F}=$ `Skywork-DPO-Formulator-7B` and $\mathrm{RM}_\mathrm{S}=$ `Skywork-DPO-Coder-7B`. We compare four configurations: (1) \textit{No RM}: the pipeline generates a single candidate at each stage without reward model guidance; (2) \textit{RM\textsubscript{F} Only}: only the formulation reward model is active (Best-of-$N$ selection at the formulation stage, single generation at the coding stage); (3) \textit{RM\textsubscript{S} Only}: only the coding reward model is active; (4) \textit{Full PARM}: both reward models are active. All RM-guided configurations use $N=32$ candidates.

Table~\ref{tab:per_stage_ablation} shows that Full PARM achieves the highest ER on all four datasets, with large gains on IndustryOR ($+13.0\%$) and NL4Opt ($+9.0\%$). Each individual RM also improves over No RM: on NL4Opt, either RM alone raises ER from $0.79$ to $0.85$ and SA from $0.39$ to $0.47$--$0.48$. The two RMs contribute through different mechanisms---on NLP4LP the coding RM dominates (SA $0.12$ vs.\ $0.08$), whereas on IndustryOR both contribute equally---and their combination on NL4Opt yields an SA gain ($+0.11$) that exceeds either alone, confirming non-redundant contributions. This ablation isolates the inference-time selector effect; the training-time cross-stage supervision is formalized separately in Section~\ref{reward_training}.

\begin{table}[!t]
    \caption{Per-Stage Reward Model Ablation (Qwen-Series, $N\!=\!32$, w/o Self-Debugging; default PARM checkpoints $\mathrm{RM}_\mathrm{F}=$ Skywork-DPO-Formulator-7B and $\mathrm{RM}_\mathrm{S}=$ Skywork-DPO-Coder-7B). \textbf{Bold}: best in column.}
    \label{tab:per_stage_ablation}
    \centering
    \resizebox{\columnwidth}{!}{%
    \begin{tabular}{lcccccccc}
        \toprule
        \multirow{2}{*}{Configuration} & \multicolumn{4}{c}{ER} & \multicolumn{4}{c}{SA} \\
        \cmidrule(lr){2-5} \cmidrule(lr){6-9}
         & NL4Opt & ComplexOR & IndustryOR & NLP4LP & NL4Opt & ComplexOR & IndustryOR & NLP4LP \\
        \midrule
        No RM         & 0.79 & 0.26 & 0.40 & 0.32 & 0.39 & 0.11 & 0.09 & 0.09 \\
        RM\textsubscript{F} Only & 0.85 & 0.26 & 0.46 & 0.35 & 0.47 & 0.11 & 0.12 & 0.08 \\
        RM\textsubscript{S} Only & 0.85 & 0.26 & 0.45 & 0.38 & 0.48 & 0.11 & 0.12 & 0.12 \\
        Full PARM     & \textbf{0.88} & \textbf{0.32} & \textbf{0.53} & \textbf{0.42} & \textbf{0.50} & 0.11 & \textbf{0.13} & \textbf{0.12} \\
        \bottomrule
    \end{tabular}%
    }
\end{table}

\textbf{Effect of Candidate Pool Size ($N$).}
A natural question is whether the benefit of DPO-trained reward models scales with the number of candidates $N$ used in Best-of-$N$ selection. To investigate this, we evaluate the full PARM pipeline (both RM\textsubscript{F} and RM\textsubscript{S} active) at $N \in \{16, 32, 48\}$ on all four datasets. As a control, we also evaluate a \textit{Random-of-$N$} baseline that generates $N$ candidates but selects one uniformly at random, without reward model guidance.

Table~\ref{tab:n_scaling} presents the results. Two clear trends emerge:

\begin{itemize}
    \item \textit{RM-guided selection improves consistently with larger $N$.} On three of four datasets, SA increases monotonically from $N\!=\!16$ to $N\!=\!48$, while ER is non-decreasing on all four datasets. On NL4Opt, SA rises from $0.47$ to $0.52$ ($+5\%$ absolute); on IndustryOR, SA rises from $0.11$ to $0.16$ ($+5\%$); on NLP4LP, SA rises from $0.11$ to $0.14$ ($+3\%$). ComplexOR (19 problems) shows a ceiling effect at SA$\,=\,0.11$, but ER increases from $0.26$ to $0.32$ at $N\!\geq\!32$.
    \item \textit{Random selection does not benefit systematically from larger $N$.} On NL4Opt, SA is $0.39$ at $N\!=\!48$, identical to the No-RM single-generation baseline. On IndustryOR, random-of-$N$ SA plateaus at $0.11$ from $N\!=\!16$ to $N\!=\!48$, while on NLP4LP it varies between $0.09$ and $0.11$ without an upward trend. This confirms that the scaling gains are attributable to the reward model's discriminative ability, not to brute-force enumeration.
\end{itemize}

These results demonstrate that DPO training produces reward models with genuine ranking capability: as the candidate pool grows, the trained RM continues to convert the additional search budget into higher-quality outputs, a property that would not hold if the reward model were merely adding noise to the selection process.

\begin{table}[!t]
    \caption{Effect of Candidate Pool Size $N$ on Full PARM Performance (Qwen-Series, w/o Self-Debugging). \textbf{Bold}: best in column.}
    \label{tab:n_scaling}
    \centering
    \resizebox{\columnwidth}{!}{%
    \begin{tabular}{llcccccccc}
        \toprule
        \multirow{2}{*}{Config} & \multirow{2}{*}{$N$} & \multicolumn{4}{c}{ER} & \multicolumn{4}{c}{SA} \\
        \cmidrule(lr){3-6} \cmidrule(lr){7-10}
        & & NL4Opt & ComplexOR & IndustryOR & NLP4LP & NL4Opt & ComplexOR & IndustryOR & NLP4LP \\
        \midrule
        No RM       & 1  & 0.79 & 0.26 & 0.40 & 0.32 & 0.39 & 0.11 & 0.09 & 0.09 \\
        \midrule
        Full PARM   & 16 & 0.88 & 0.26 & 0.51 & 0.38 & 0.47 & 0.11 & 0.11 & 0.11 \\
        Full PARM   & 32 & 0.88 & 0.32 & 0.53 & 0.42 & 0.50 & 0.11 & 0.13 & 0.12 \\
        Full PARM   & 48 & \textbf{0.91} & \textbf{0.32} & \textbf{0.55} & \textbf{0.46} & \textbf{0.52} & 0.11 & \textbf{0.16} & \textbf{0.14} \\
        \midrule
        Random-of-$N$ & 16 & 0.80 & 0.26 & 0.45 & 0.32 & 0.39 & 0.11 & 0.11 & 0.11 \\
        Random-of-$N$ & 48 & 0.79 & 0.26 & 0.42 & 0.35 & 0.39 & 0.11 & 0.11 & 0.09 \\
        \bottomrule
    \end{tabular}%
    }
\end{table}

\begin{table*}[!t]
\caption{Comparison of Execution Rate (ER) and Solving Accuracy (SA) on PARM with Various Reward and Expert Models (Without Self-Debugging). Bold values indicate the best performance in each column.}
\label{main_experiment}
\centering
\resizebox{\textwidth}{!}{
\begin{tabular}{cccccccccccc}
\toprule
 & \multirow{2}{*}{Method} & \multirow{2}{*}{Math RM} & \multirow{2}{*}{Code RM} & \multicolumn{4}{c}{ER} & \multicolumn{4}{c}{SA} \\
\cmidrule(lr){5-8} \cmidrule(lr){9-12}
 & & & & IndustryOR & ComplexOR & NL4Opt & NLP4LP & IndustryOR & ComplexOR & NL4Opt & NLP4LP \\
\midrule
\multirow{5}{*}{Qwen-Series} 
    & Sampling Decoding & - & - & 0.40 & 0.26 & 0.79 & 0.32 & 0.09 & 0.11 & 0.39 & 0.09 \\
    \cmidrule(lr){2-12}
    & Reward Model & Skywork & Skywork & 0.50 & 0.32 & 0.92 & 0.41 & 0.11 & \textbf{0.16} & 0.44 & 0.11 \\
    & Reward Model & Qwen-RM & Skywork & 0.51 & 0.31 & 0.86 & \textbf{0.48} & 0.13 & 0.05 & 0.49 & 0.08 \\
    & PARM (ours) & Skywork(DPO) & Skywork(DPO) & 0.53 & 0.32 & 0.88 & 0.42 & 0.13 & 0.11 & \textbf{0.50} & \textbf{0.12} \\
    & PARM (ours) & Qwen-RM(DPO) & Skywork(DPO) & \textbf{0.54} & \textbf{0.37} & \textbf{0.94} & 0.45 & \textbf{0.16} & 0.11 & 0.49 & 0.13 \\
\midrule
\multirow{5}{*}{DeepSeek-Series} 
    & Sampling Decoding & - & - & 0.07 & 0.05 & 0.27 & 0.03 & 0.01 & 0.00 & 0.13 & 0.02 \\
    \cmidrule(lr){2-12}
    & Reward Model & Skywork & Skywork & 0.17 & 0.11 & 0.58 & 0.08 & \textbf{0.13} & 0.05 & 0.28 & 0.05 \\
    & Reward Model & Qwen-RM & Skywork & 0.14 & 0.05 & 0.45 & 0.08 & 0.05 & 0.00 & 0.21 & 0.05 \\
    & PARM (ours) & Skywork(DPO) &  Skywork(DPO)  & \textbf{0.25} & 0.11 & \textbf{0.59} & 0.08 & 0.07 & 0.05 & \textbf{0.30} & 0.05 \\
    & PARM (ours) & Qwen-RM(DPO) &  Skywork(DPO) & 0.15 & \textbf{0.16} & 0.52 & 0.08 & 0.08 & \textbf{0.11} & 0.24 & 0.05 \\        
\bottomrule
\end{tabular}%
}
\end{table*}

\begin{table*}[!t]
\caption{Comparison of Execution Rate (ER) and Solving Accuracy (SA) on PARM with Various Reward and Expert Models, Including Self-Debugging (1 Iteration). Bold values indicate the best performance in each column.}
\label{self_debug}
\centering
\resizebox{\textwidth}{!}{%
\begin{tabular}{lcccccccccccc}
\toprule
 & Method & Math RM & Code RM & \multicolumn{4}{c}{ER} & \multicolumn{4}{c}{SA} \\
\cmidrule(lr){5-8} \cmidrule(lr){9-12}
 & & & & IndustryOR & ComplexOR & NL4Opt & NLP4LP & IndustryOR & ComplexOR & NL4Opt & NLP4LP \\
\midrule
\multirow{5}{*}{Qwen-Series}
    & Sampling & - & - & 0.56 & 0.32 & 0.93 & 0.40 & 0.09 & 0.11 & \textbf{0.53} & 0.12 \\
    \cmidrule(lr){2-12}
    & Reward Model & Skywork & Skywork & 0.66 & 0.32 & 0.97 & 0.52 & \textbf{0.15} & 0.16 & 0.51 & 0.11 \\
    & Reward Model & Qwen-RM & Skywork & 0.60 & 0.42 & 0.96 & 0.55 & 0.14 & 0.16 & \textbf{0.53} & 0.11 \\
    \cmidrule(lr){2-12}
    & PARM (ours) & Skywork(DPO) & Skywork(DPO) & \textbf{0.70} & 0.47 & 0.97 & \textbf{0.56} & \textbf{0.15} & 0.17 & 0.52 & \textbf{0.15} \\
    & PARM (ours) & Qwen-RM(DPO) & Skywork(DPO) & 0.64 & \textbf{0.57} & \textbf{0.98} & 0.51 & 0.14 & \textbf{0.21} & 0.50 & 0.13 \\
\bottomrule
\end{tabular}%
}
\end{table*}

\subsubsection{\textbf{Evaluation on Reward Model}}
To evaluate the impact of reward models within the PARM pipeline, we systematically investigate both the construction of training data and the effect of different reward model configurations on downstream optimization performance.

\textbf{Automated Construction of Preference Data.}
Unlike prior works that require costly manual annotation, we employ an automated process to generate preference pairs for reward model training. Specifically, we run the pipeline (without reward models) on randomly sampled problems from the Mamo dataset, divided into Easy and Complex subsets. For each problem, multiple formulations and codes are generated; we use the task-level verification criterion described in Section~\ref{reward_training} to label preference data. This yields two types of pairs as required by Direct Preference Optimization (DPO): [problem, formulation\textsubscript{chosen}, formulation\textsubscript{rejected}] and [formulation, code\textsubscript{chosen}, code\textsubscript{rejected}]. The procedure is fully automatic and tractable in the studied two-stage setting.

Table~\ref{reward_model_datasets_num} summarizes the number of preference data pairs obtained from different sampling settings. We observe that more complex problems yield fewer correct solutions, but contribute richer and harder supervision signals for reward model training.

\begin{table}[!t]
    \caption{Number of Preference Data Pairs Constructed from the Mamo Dataset for DPO Training}
    \label{reward_model_datasets_num}
    \centering
    \resizebox{\columnwidth}{!}{%
    \begin{tabular}{lcccc}
        \toprule
        \textbf{Subset} & \textbf{Sample Size} & \textbf{\# (p, f\textsuperscript{+}, f\textsuperscript{--})} & \textbf{\# (f, s\textsuperscript{+}, s\textsuperscript{--})} \\
        \midrule
        MamoEasy (50)    & 50     & 752     & 49,927  \\
        MamoComplex (50) & 50     & 2,542   & 40,806  \\
        MamoEasy (100)   & 100    & 887     & 52,906 \\
        MamoComplex (100)& 100    & 4,377   & 59,549  \\
        \bottomrule
    \end{tabular}%
    }
\end{table}

\textbf{Influence on Pipeline Performance.}
To directly assess the practical value of reward model training, we compare pipeline execution rates (ER) using different reward model combinations and data regimes, as reported in Table~\ref{statistic_RM}. The findings are as follows:
\begin{itemize}
    \item Reward models trained with larger and more complex datasets lead to higher ER on challenging benchmarks.
    \item Matching Math RM and Code RM from different data regimes (e.g., Math RM from Complex, Code RM from Easy) can further enhance performance, indicating complementary effects.
    \item The improvements are most pronounced on datasets with higher complexity, validating the benefit of hard supervision.
\end{itemize}

\begin{table*}[ht!]
    \centering
    \caption{Execution Rate (ER) of Different Reward Model Combinations on Four Datasets (w/o self-debugging)}
    \begin{tabular}{ll|llll}
    \hline
    \multicolumn{1}{l|}{Math RM} & Code RM & \multicolumn{4}{c}{ER (w/o self-debugging)} \\
    \hline
    \multicolumn{2}{c|}{DPO on Skywork-RM} & IndustryOR & ComplexOR & NL4Opt & NLP4LP \\
    \hline
    MamoEasy(50)     & MamoEasy(50)    & 0.49 & 0.29 & 0.83 & 0.38 \\
    MamoEasy(100)    & MamoEasy(50)    & 0.53 & 0.32 & 0.89 & 0.40 \\
    MamoComplex(50)  & MamoComplex(50) & 0.53 & 0.30 & 0.90 & 0.45 \\
    MamoComplex(100) & MamoComplex(50) & 0.53 & 0.32 & 0.90 & 0.46 \\
    MamoComplex(100) & MamoEasy(50)    & \textbf{0.54} & \textbf{0.42} & \textbf{0.92} & \textbf{0.46} \\
    \hline
    \multicolumn{2}{c|}{DPO on Qwen-RM} & IndustryOR & ComplexOR & NL4Opt & NLP4LP \\
    \hline
    MamoEasy(50)     & MamoEasy(50)    & 0.49 & 0.26 & 0.79 & 0.42 \\
    MamoEasy(100)    & MamoEasy(50)    & 0.52 & 0.32 & 0.90 & 0.45 \\
    MamoComplex(50)  & MamoComplex(50) & 0.53 & 0.29 & 0.86 & 0.45 \\
    MamoComplex(100) & MamoComplex(50) & 0.54 & 0.32 & 0.86 & 0.45 \\
    MamoComplex(100) & MamoEasy(50)    & 0.53 & 0.32 & 0.90 & 0.45 \\
    \hline
    \end{tabular}
    \label{statistic_RM}
\end{table*}

\begin{table}[!t]
    \caption{Pairwise Preference Accuracy (\texttt{eval\_rewards/accuracies}) of Math and Code Reward Models on the Evaluation Set After DPO Training}
    \label{reward_model_training}
    \centering
    \resizebox{\columnwidth}{!}{%
    \begin{tabular}{lcc|c}
        \toprule
        \textbf{Dataset} & \multicolumn{2}{c|}{\textbf{Math RM}} & \textbf{Code RM} \\
        \cmidrule(lr){2-3} \cmidrule(lr){4-4}
         & Skywork & Qwen-RM & Skywork \\
        \midrule
        MamoEasy (50)     & 0.5625 & 0.5125 & \textbf{0.8769} \\
        MamoComplex (50)  & 0.6351 & \textbf{0.6422} & 0.7647 \\
        MamoEasy (100)    & 0.6573 & 0.6042 & 0.8563 \\
        MamoComplex (100) & \textbf{0.6924} & 0.6295 & 0.8395 \\
        \bottomrule
    \end{tabular}%
    }
\end{table}

\textbf{Reward Model Training and Selection.}
We use Direct Preference Optimization (DPO) to fine-tune both math and code reward models, leveraging preference pairs automatically constructed from the Mamo dataset. Table~\ref{reward_model_datasets_num} details the number of preference pairs generated for each data subset, illustrating that larger sample sizes and more complex problems yield substantially more training pairs.

To evaluate reward model performance, we report pairwise preference accuracy (\texttt{eval\_rewards/accuracies}) on a held-out evaluation set, as shown in Table~\ref{reward_model_training}. This metric reflects the proportion of evaluation preference pairs where the reward model correctly assigns a higher score to the preferred (chosen) output, providing a direct measure of the model's discrimination ability after DPO training. The results reveal several trends:
\begin{itemize}
    \item Increasing the number of preference pairs, particularly for more challenging subsets, generally improves the pairwise accuracy of reward models. However, we also observe minor fluctuations; for example, the Qwen-RM math reward model achieves slightly higher accuracy on MamoComplex (50) compared to MamoComplex (100), despite the greater number of training pairs. This may be due to increased data diversity, harder evaluation samples, or inherent variance in small-scale experiments.
    \item Both Skywork and Qwen-RM serve as effective base models for reward learning, with the optimal configuration varying depending on the specific task and dataset characteristics.
\end{itemize}

Based on these results, we select the best-performing Math RM and Code RM (according to pairwise preference accuracy) as the default reward models for our pipeline.

The DPO training curves show stable convergence: evaluation loss decreases monotonically and pairwise preference accuracy rises from approximately 0.59 to over 0.80, consistent with the final values reported in Table~\ref{reward_model_training}.

\textbf{Conclusion.}

This set of experiments demonstrates that (1) automated collection of preference data is effective for reward model training in the studied two-stage pipeline, (2) increasing the size and complexity of training data significantly benefits reward model accuracy and downstream pipeline performance, and (3) careful pairing of reward models from different data regimes can further optimize results. Overall, our approach supports the flexible and robust integration of reward models in the evaluated setting.

\subsection{Supplementary Generalizability Check on Math Reasoning (GSM8K)}
\label{sec:gsm8k}
To demonstrate that PARM can extend beyond the optimization domain, we add a supplementary experiment on a fundamentally different task: grade-school math reasoning on the GSM8K benchmark~\cite{cobbe2021training}. This experiment is intended as a targeted cross-domain validation rather than a second main benchmark suite. It addresses two key concerns: (1) whether PARM's methodology transfers when the task semantics, intermediate representation, and evaluation protocol change, and (2) whether stage-specific reward models provide meaningful gains in a domain where the underlying generator is already strong.

\subsubsection{\textbf{Task and Pipeline Design}}
We instantiate a two-stage PARM pipeline for GSM8K (1{,}319 test problems):
\begin{itemize}
    \item \textbf{Stage~1 (Planning):} Decompose a math word problem into a structured solution plan (equations, reasoning steps) using Qwen2.5-Math-7B-Instruct.
    \item \textbf{Stage~2 (Solving):} Translate the plan into an executable Python program whose output is the numerical answer using Qwen2.5-Coder-7B-Instruct.
\end{itemize}
This pipeline differs from the optimization pipeline in several important respects: (a) the domain is elementary mathematics rather than operations research; (b) the intermediate representation is a natural-language solution plan rather than a formal optimization formulation; and (c) correctness is determined by exact numerical match rather than feasibility and optimality checking. These differences provide a targeted test of cross-domain transfer.

\subsubsection{\textbf{Training Data and Reward Models}}
Following the same PARM methodology (Section~\ref{reward_training}), we constructed DPO training data from the full GSM8K training set (7{,}473 problems). For each problem, 10 plans were sampled; for each plan, 10 code candidates were generated and executed. Cross-stage credit assignment produced 9{,}318 plan preference pairs and 46{,}582 solver preference pairs. Two stage-specific LoRA adapters were trained on Skywork-o1-Open-PRM-Qwen-2.5-7B via DPO, using the same overall hyperparameter regime as the optimization experiments (LoRA rank $= 128$, $\alpha = 64$, learning rate $= 5.0 \times 10^{-7}$), with 3 training epochs for GSM8K.

\subsubsection{\textbf{Results}}
We evaluate the same ablation matrix as the optimization experiments, using Best-of-$N$ selection with $N = 32$ candidates per stage at inference time. Table~\ref{tab:gsm8k_ablation} presents the results.

\begin{table}[!t]
    \caption{Supplementary Per-Stage Reward Model Ablation on GSM8K (Qwen2.5-Math-7B-Instruct $\rightarrow$ Qwen2.5-Coder-7B-Instruct, $N\!=\!32$). \textbf{Bold}: best in column.}
    \label{tab:gsm8k_ablation}
    \centering
    \begin{tabular}{lcccc}
        \toprule
        Configuration & Plan RM & Solver RM & ER & SA \\
        \midrule
        Random-of-$N$          & \textemdash & \textemdash & 0.984 & 0.923 \\
        No RM                  & \textemdash & \textemdash & 0.986 & 0.927 \\
        RM\textsubscript{F} Only & \checkmark & \textemdash & 0.993 & 0.945 \\
        RM\textsubscript{S} Only & \textemdash & \checkmark & 0.993 & 0.931 \\
        \textbf{Full PARM}     & \checkmark & \checkmark & \textbf{1.000} & \textbf{0.963} \\
        \bottomrule
    \end{tabular}
\end{table}

Key findings: (1)~Full PARM achieves 96.3\% SA, a $+3.6$~pp improvement over the No-RM baseline (92.7\%) and $+4.0$~pp over random selection, confirming effectiveness in math reasoning. (2)~Both RMs contribute non-redundantly: the Plan RM alone raises SA to 94.5\% ($+1.8$~pp) and the Solver RM to 93.1\% ($+0.4$~pp). Notably, the Plan RM contributes more strongly here than in optimization, where the Code RM often dominated---intuitively, plan quality is the primary bottleneck for math word problems, whereas code generation is relatively straightforward once a correct plan is given to a code-specialized solver. This validates that stage-specific reward models capture domain-specific bottlenecks. (3)~Full PARM reaches 100\% ER, yielding executable programs for all test problems.

\subsubsection{\textbf{Cross-Domain Implications}}
The GSM8K experiment provides supplementary evidence that PARM transfers to a fundamentally different domain. The task semantics (grade-school math vs. operations research), intermediate representation (solution plan vs. formal formulation), and evaluation protocol (exact numerical match vs. feasibility and optimality checking) all differ from the optimization experiments, yet PARM still improves both ER and SA with the same pattern of complementary stage-specific contributions. We present GSM8K as a targeted proof-of-concept; broader validation across additional non-optimization task families remains future work.

\section{Discussion}

In this work, we show that pipeline frameworks guided by automatically trained reward models are a promising paradigm for complex reasoning pipelines. In the evaluated two-stage setting, our pipeline mechanism enables modular integration of lightweight expert models and leverages reward signals derived directly from execution feedback rather than manual annotation. This design yields robust and reproducible workflows that can surpass monolithic large-model baselines in the studied tasks, while broader scalability to richer pipeline structures remains to be established.

A key innovation of our approach lies in the automated construction of preference datasets for reward model training. By collecting supervision signals directly from pipeline execution without human labeling, we enable automatic reward-model adaptation in the studied setting. This also keeps the framework modular: new expert models or reward objectives can, in principle, be incorporated without changing the core training recipe, although the cost of data construction can grow substantially as additional true stages are introduced.

In terms of generalizability, our main results on optimization benchmarks, together with the supplementary GSM8K experiment, suggest that the pipeline framework can transfer beyond a single task domain. However, broader validation across additional domains—such as open-domain reasoning or multi-modal tasks—remains an open direction for future investigation. We also note that while our reward models operate without manually annotated data, incorporating a small amount of high-quality human feedback (especially for formulation tasks) could further improve overall performance and reliability.

Looking ahead, future work may explore lightweight or online reward model adaptation, hybrid human-in-the-loop supervision, more interpretable reward signals, and integration of advanced search strategies such as Monte Carlo Tree Search.

\textbf{Scalability to Multi-Stage Pipelines.}
A natural question is whether PARM scales beyond the two-stage setting evaluated in our experiments. We discuss this from three angles.

\emph{Inference-time tractability.}
Exhaustively enumerating all stage combinations in a $k$-stage pipeline with $N$ candidates per stage incurs $O(N^k)$ joint evaluations. PARM avoids this via greedy stage-wise selection: each reward model scores its $N$ candidates and forwards only the best to the next stage, yielding $O(k \cdot N)$ RM operations. This removes the combinatorial blowup at selection time, although training-data construction still grows with downstream rollouts.

\emph{Cross-stage credit assignment and data construction.}
The main challenge of extending to more stages is the sparsity and cost of early-stage supervision. In our two-stage pipeline, a formulation is labeled positive only if at least one downstream solution is verified; in a $k$-stage pipeline, positive labels for early stages depend on successful continuations through the \emph{entire remaining pipeline}, making them rarer and more expensive. Table~\ref{reward_model_datasets_num} shows an order-of-magnitude imbalance between formulation preference pairs (752--4,377) and code preference pairs (40,806--59,549), which would intensify with additional stages. Addressing this---e.g., through intermediate reward shaping or partial credit assignment---is an important direction for future work.

\textbf{Broader Applicability.}
Although this paper instantiates PARM on mathematical optimization pipelines, the framework applies more broadly whenever a task decomposes into multiple LLM-mediated stages and end-to-end feedback is objectively computable (e.g., test pass/fail, simulation cost, answer correctness under a verifier). Representative scenarios include automated programming (requirement understanding $\rightarrow$ code synthesis $\rightarrow$ test execution), supply-chain planning with simulator-backed evaluation, and closed-domain RAG pipelines where retrieval quality can be judged by end-to-end answer correctness. When such feedback is unavailable or highly subjective, PARM becomes less suitable without additional supervision.

\textbf{Limitations.}
Despite its promising results, this study has several limitations. While we now provide supplementary evidence beyond optimization through GSM8K, broader validation across additional domains (e.g., open-domain reasoning, multi-modal tasks) and in pipelines with three or more true stages remains an open challenge---particularly due to cross-stage credit-assignment sparsity, as discussed above. Furthermore, although our method eliminates the need for manual annotation, a small amount of high-quality labeled data might further enhance reward model performance, particularly for subtle or domain-specific tasks. We leave these aspects for future investigation.

\section{Conclusion}
In this paper, we introduced PARM, a modular pipeline framework for complex optimization tasks that combines lightweight expert models with reward models trained through automated preference data collection. By decomposing problems into sub-tasks and integrating specialized components such as the Formulator and Coder, PARM enables flexible composition and iterative refinement of solutions. Our framework leverages reward models that are trained entirely without manual annotation, using real execution-based feedback to guide the selection of high-quality intermediate outputs.

Extensive experiments across multiple benchmarks demonstrate that PARM not only outperforms direct large-model baselines in solving accuracy and execution rate, but also exhibits robustness and effectiveness in the evaluated two-stage setting. The inclusion of reward models and a self-debugging mechanism further boosts performance, confirming the value of reward-guided, iterative optimization within a modular architecture. Importantly, our automated data collection process shows that reward-model adaptation can be carried out without manual annotation in this setting.

Our work highlights the potential of pipeline-based reasoning frameworks that are both extensible and interpretable. For future research, we aim to generalize PARM to broader domains beyond code generation, investigate advanced reward modeling strategies, and further explore the integration of human feedback and advanced search techniques. We believe that the pipeline paradigm, coupled with automated reward learning, is a promising direction for more robust and transparent problem solving, while broader validation across additional domains and true 3+ stage pipelines remains future work.


\bibliographystyle{IEEEtran}
\bibliography{references}{}

\section{Biography Section}
\begin{IEEEbiography}
[{\includegraphics[width=1in,height=1.25in,clip,keepaspectratio]{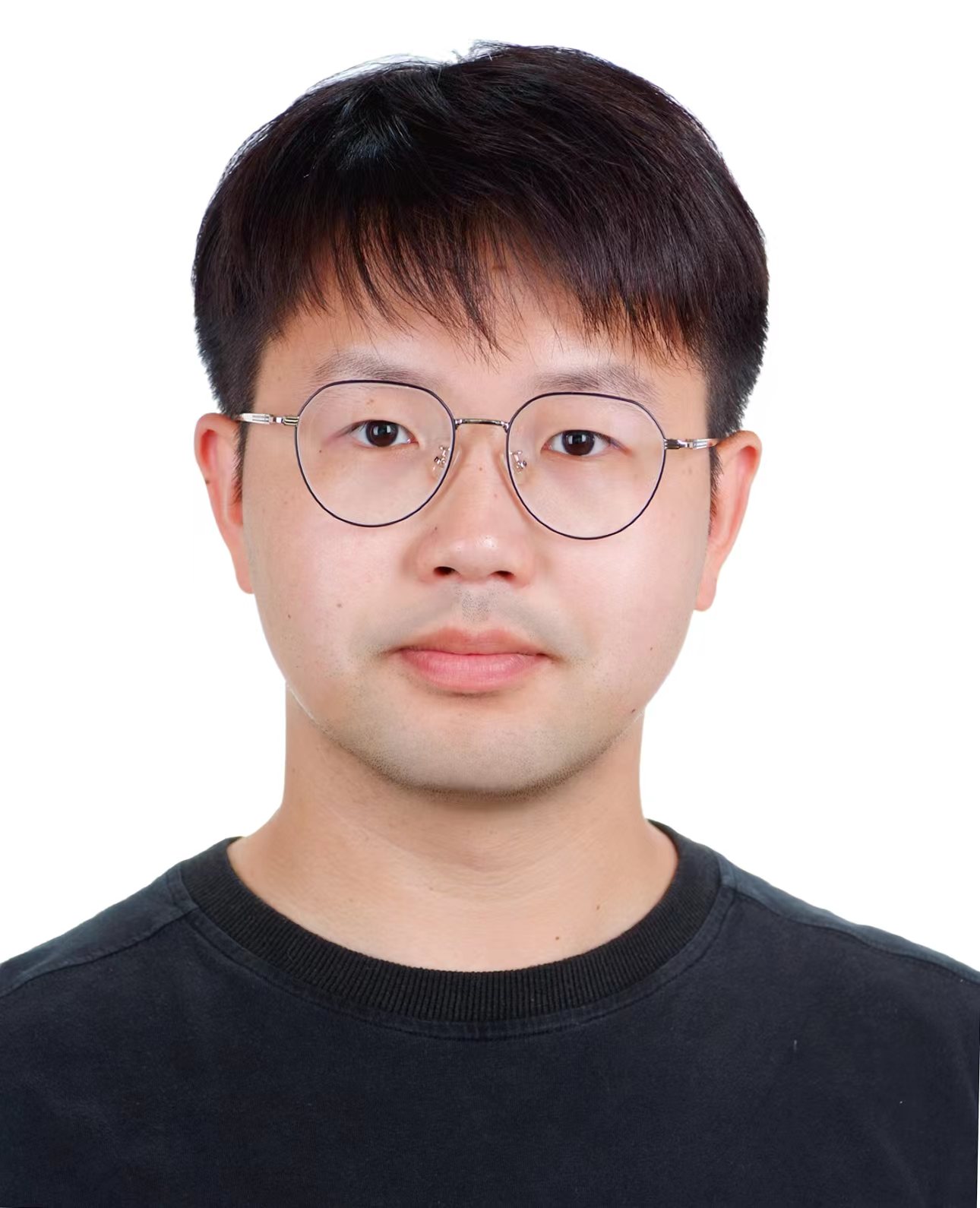}}]{Xingyu Fan} received the B.Sc. degree in software engineering from University of Electronic Science and Technology of China in 2022. Since 2022, he has been working toward the Ph.D. degree with the Department of Computer Science and Engineering, The Chinese University of Hong Kong, Hong Kong. His research interests include Reasoning Large Language Model, AI4Biology, Graph Learning. 
\end{IEEEbiography}

\begin{IEEEbiography}[{\includegraphics[width=1in,height=1.25in,clip,keepaspectratio]{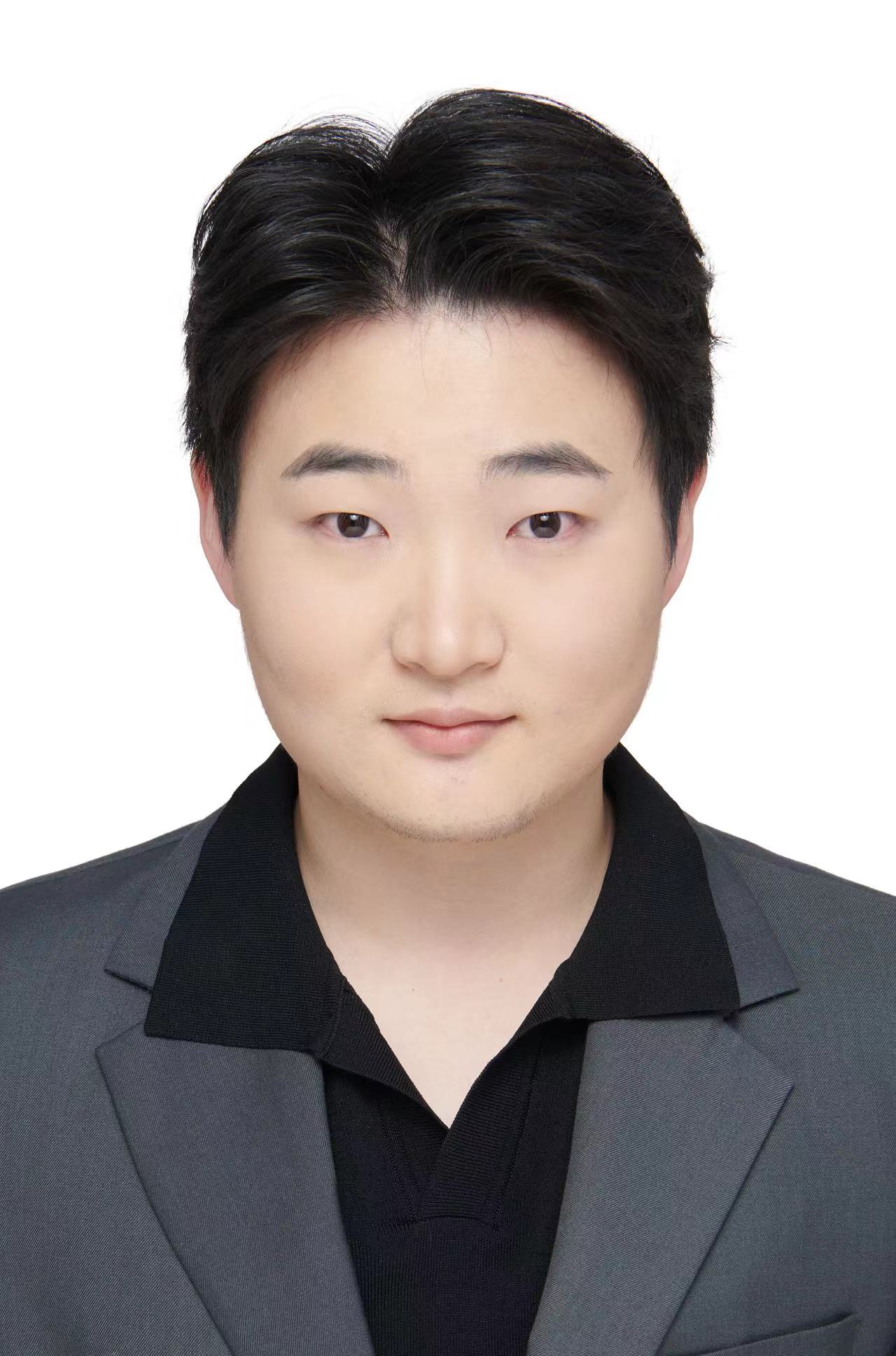}}]{Wei Shao}
 Dr. Wei Shao received his Ph.D. degree (2024) in Computer Science at City University of Hong Kong under the supervision of Prof. Linqi Song, B.S. (2020) from Peking University. His research interests includes: Machine Learning (Reinforcement Learning), Large Language Model Inferece Acceleration.
\end{IEEEbiography}

\begin{IEEEbiography}[{\includegraphics[width=1in,height=1.25in,clip,keepaspectratio]{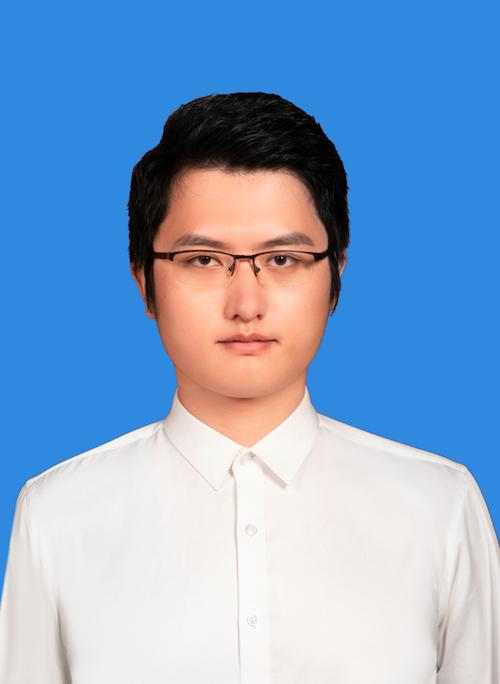}}]{Jiacheng Liu}
(Member, IEEE) received the Ph.D. degree in 2022 from Shanghai Jiao Tong University (SJTU), Shanghai, China. He is currently a post-doctoral fellow at Hong Kong University of Science and Technology (HKUST). Prior to this, he worked as a post-doctoral fellow at the Chinese University of Hong Kong (CUHK). His research interests include Large Language Model, AI for Science. 
\end{IEEEbiography}
\vspace{11pt}

\begin{IEEEbiography}[{\includegraphics[width=1in,height=1.25in,clip,keepaspectratio]{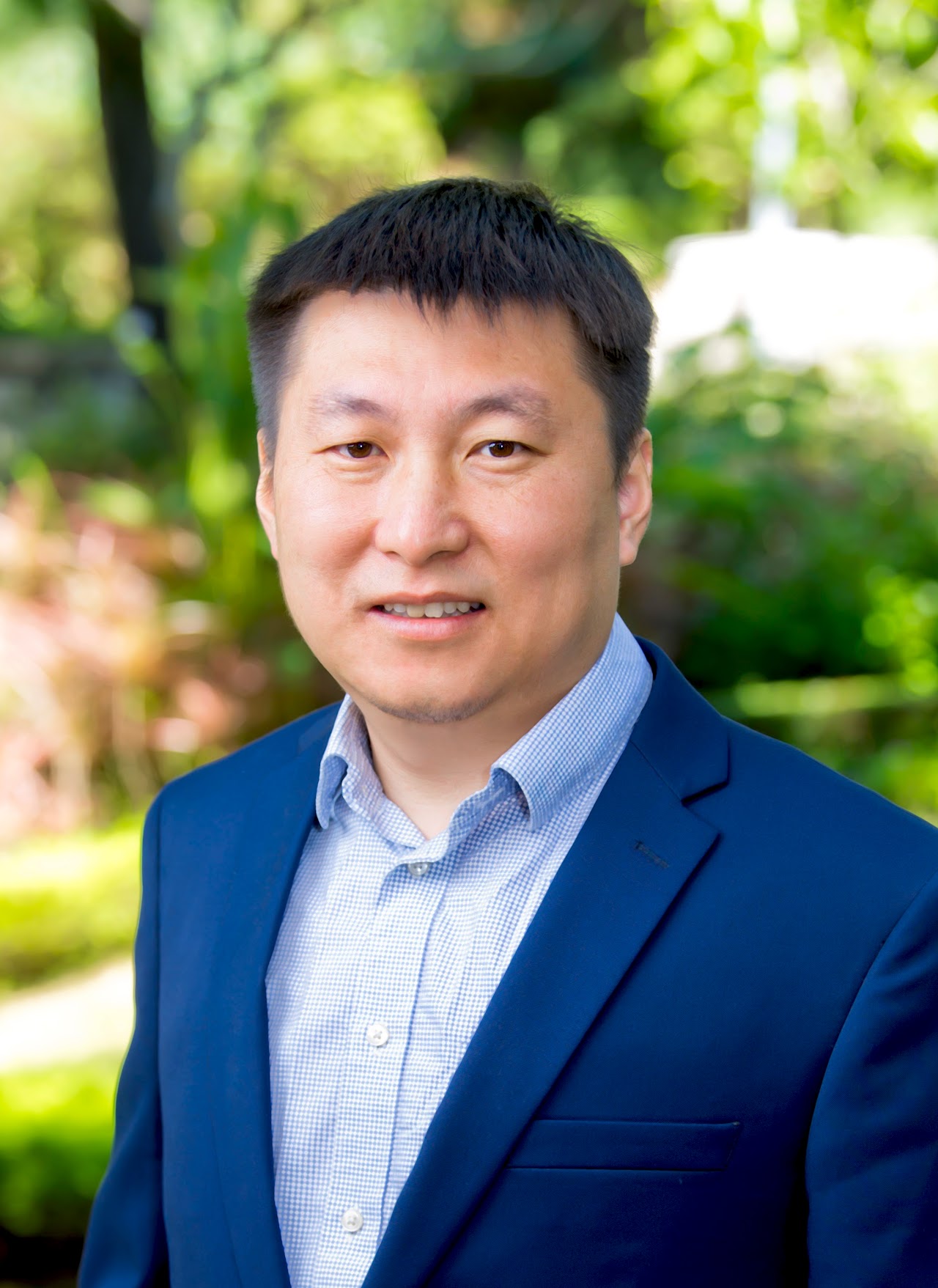}}]{Linqi Song}
(Member, IEEE) Dr. Linqi Song is an Associate Professor in the Department of Computer Science at the City University of Hong Kong. Prior to that he was a Postdoctoral Scholar in the Electrical and Computer Engineering Department at UCLA. He received his Ph.D. degree in Electrical Engineering at UCLA under the supervision of Prof. Christina Fragouli (ARNI lab), B.S. and M.S. from Tsinghua University. His research interests includes: Information Theory, Data Science, Machine Learning (Reinforcement Learning), NLP, Algorithms, Recommender Systems.
\end{IEEEbiography}

\begin{IEEEbiography}[{\includegraphics[width=1in,height=1.25in,clip,keepaspectratio]{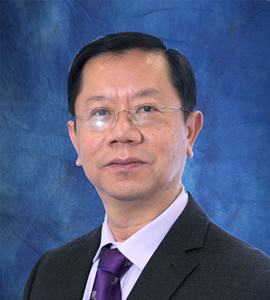}}]{Pheng Ann Heng}
 is Choh-Ming Li Professor of Computer Science and Engineering at The Chinese University of Hong Kong. He is the Director of the Institute of Medical Intelligence and XR. His research interests include AI/XR for medical and scientific applications, visualization, graphics, human-computer interaction and computer vision. He is recognized as a Highly Cited Researcher by Clarivate in 2024 and also recognized with the Research.com Computer Science in China Leader Award since 2023. According to Google Scholar, his publications have been cited over 60,000 times, with an H-index of 117.
\end{IEEEbiography}
\vspace{11pt}

\vfill

\end{document}